\documentclass[journal]{IEEEtran}

\usepackage{times}
\usepackage{epsfig}
\usepackage{graphicx}
\usepackage{amsmath}
\usepackage{amssymb}
\usepackage{cite}
\usepackage{subfigure}
\usepackage{booktabs}
\usepackage{bm}
\usepackage{enumitem}
\usepackage{dblfloatfix}
\usepackage{url}

\begin{document}

\title{Background-Aware 3D Point Cloud Segmentation with Dynamic Point Feature Aggregation}

\author{Jiajing~Chen\textsuperscript{*},
        Burak~Kakillioglu\textsuperscript{*},
        and~Senem~Velipasalar,~\IEEEmembership{Senior Member,~IEEE}
\thanks{Authors are with the Department
of Electrical Engineering and Computer Science, Syracuse University, Syracuse
NY, 13244 USA e-mail: \{jchen152,bkakilli,svelipas@syr.edu\}.}
\thanks{(*) Authors contributed equally.}
}


\maketitle

\begin{abstract}
With the proliferation of Lidar sensors and 3D vision cameras, 3D point cloud analysis has attracted significant attention in recent years. After the success of the pioneer work PointNet, deep learning-based methods have been increasingly applied to various tasks, including 3D point cloud segmentation and 3D object classification. In this paper, we propose a novel 3D point cloud learning network, referred to as Dynamic Point Feature Aggregation Network (DPFA-Net), by selectively performing the neighborhood feature aggregation with dynamic pooling and an attention mechanism. DPFA-Net has two variants for semantic segmentation and classification of 3D point clouds. As the core module of the DPFA-Net, we propose a Feature Aggregation layer, in which features of the dynamic neighborhood of each point are aggregated via a self-attention mechanism. In contrast to other segmentation models, which aggregate features from fixed neighborhoods, our approach can aggregate features from different neighbors in different layers providing a more selective and broader view to the query points, and focusing more on the relevant features in a local neighborhood. In addition, to further improve the performance of the proposed semantic segmentation model, we present two novel approaches, namely Two-Stage BF-Net and BF-Regularization to exploit the background-foreground information. Experimental results show that the proposed DPFA-Net achieves the state-of-the-art overall accuracy score for semantic segmentation on the S3DIS dataset, and provides a consistently satisfactory performance across different tasks of semantic segmentation, part segmentation, and 3D object classification. It is also computationally more efficient compared to other methods.
\end{abstract}


\section{Introduction}
\IEEEPARstart{T}{hree} dimensional (3D) point cloud data is a set of points in the 3D space, which can be gathered by special vision sensors, such as Lidar sensors and stereo vision cameras. With the ever-increasing availability of 3D vision sensors in the recent years, more attention has been paid to the 3D point cloud processing, and several methods have been proposed to extract useful information from point cloud data~\cite{qi2017pointnet,qi2017pointnet++,wang2019dynamic,hu2020randla,chen2020hierarchical}. Utilization of 3D point clouds has already been adopted in a wide range of applications, including robotics, intelligent vehicles, autonomous mapping and navigation. However, different from a regular 2D image, a point cloud is an unstructured data, which makes traditional Convolution Neural Networks (CNN) not readily applicable to point cloud analysis. PointNet~\cite{qi2017pointnet} is a pioneer work, which has shown that a symmetric function, such as maxpooling, can extract permutation invariant features from set inputs, and the authors proposed an end-to-end deep learning-based model for 3D point cloud classification and semantic segmentation.~Following PointNet, many other works have been proposed, adopting the similar idea for various problems involving point clouds \cite{qi2017pointnet++,wang2019dynamic,hu2020randla,zhao2019pointweb,liu2019point,chen2020hierarchical}.\par

Point cloud segmentation deals with automated point-wise labeling of point cloud data, which is a non-trivial task in 3D data understanding. Although it is a commonly used term, the definition of segmentation can vary from one context to another. For example, while the segmentation can mean partitioning planar segments, located on different planes, in one application, it can mean segmenting point clusters, which are distant enough from each other based on a certain criteria, in another application. In this paper, we focus on the semantic segmentation of 3D point clouds, wherein the points are categorized into different semantic classes, which are especially critical for human-machine interaction. Semantically segmented regions streamline the assignment of many tasks to autonomous agents.

A desirable network architecture for semantic point cloud segmentation in particular, and for point cloud learning in general, is the one that can effectively capture both the global information from the point cloud and the local information from sub-regions. There have been several studies that account for capturing the local information using different approaches~\cite{qi2017pointnet++,zhang2019shellnet,zhao2019pointweb,komarichev2019cnn}. In this paper, we present a novel approach and a module for a careful local neighborhood feature aggregation in point cloud learning together with two network structures for semantic segmentation and object classification. In contrast to the majority of previous works, which only gather point features from a fixed set of neighbors in each layer, we demonstrate that aggregating the higher dimensional features with the raw positional information of a dynamic set of neighbors via attention improves the final prediction result. Our proposed method uses a self-attention mechanism for selective contribution of neighbors in a local query region, by giving more weight/significance to important neighbor features and ignoring the irrelevant ones. Moreover, we also present an auxiliary methodology to further boost the performance of the semantic segmentation models. More specifically, we have observed that there is a significant imbalance between the number of background-like points and foreground-like points in a 3D point cloud scene. In our study, we show that exploiting this fact can have a positive impact on the overall accuracy of the semantic segmentation model. Hence, we present two different Background-Foreground (BF) exploitation strategies in this paper. The main contributions of this paper include the following:
\begin{itemize}
    \item A novel, attention-based and computationally efficient network, referred to as the Dynamic Point Feature Aggregation network (DPFA-Net), for local feature aggregation in point cloud learning. 
    \item 3D semantic segmentation and 3D object classification networks employing the DPFA-Net architecture.
    \item Two novel approaches for exploitation of Background-Foreground information to improve the overall semantic segmentation performance.
    \item Extensive set of experiments on multiple benchmark datasets for point cloud segmentation and classification tasks showing the state-of-the-art performance of the proposed approach in the overall accuracy for semantic segmentation.
    \item Results showing that our proposed method provides significant advantage over the compared works in terms of processing speed.
    \item The source code is provided as supplementary material, and will be made available at \url{https://github.com/jiajingchen113322/DPFA_Net}
    
\end{itemize}

The rest of this paper is organized as follows: Related work is reviewed in Section~\ref{sec:related-work}. Our proposed  DPFA-Net model, and the semantic segmentation and object classification networks are described in detail in Section~\ref{sec:dynamic-point-feature-attention}. The two novel approaches we propose for the exploitation of Background-Foreground information in semantic segmentation are presented in Section~\ref{sec:bg-segmentation}. The experimental evaluation and discussion are presented in Section~\ref{sec:experimental-results}, and the paper is concluded in Section~\ref{sec:conclusion}.


\section{Related Work}
\label{sec:related-work}


Over the last decade, CNN-based models have been proposed for solving various tasks in 2D domain, such as image classification~\cite{he2016deep,simonyan2014very}, object detection~\cite{redmon2016you,ren2015faster}, and segmentation~\cite{long2015fully,zhao2017pyramid,ronneberger2015u}. Following these developments, deep learning-based methods have emerged for 3D computer vision applications as well. Depth image-based methods, often referred to as 2.5D approaches, employed 2D depth maps with or without the corresponding RGB channels \cite{alexandre20163d, lv2017modality}. A multi-view approach was followed in \cite{su15mvcnn, johns2016pairwise}, which infers the 3D structure from different viewpoints. Panoramic representations of 2D images were used for 3D object classification in \cite{sfikas2017exploiting}. Taking a step forward towards true 3D processing, 3D binary volumetric grids (a.k.a 3D voxel grids) were used in   \cite{wu20153d,maturana2015voxnet,qi2016volumetric,song2016deepsliding,kakillioglu20203d} for 3D object classification and detection. A 3D voxel grid is a quantized version of point clouds or sampled mesh models. 

Despite the success of CNNs on 2D images, a traditional CNN is not readily applicable to raw 3D point clouds. In contrast to a 2D image or a 3D voxel grid, which are represented as a regular grid array, 3D point clouds are unstructured data composed of unordered points.  PointNet~\cite{qi2017pointnet} and DeepSets~\cite{zaheer2017deep} are two milestone works in point cloud processing with deep learning, and introduced similar ideas to overcome permutation invariance (order of the points) problem. More specifically, they proposed the use of multi-layer perceptron (MLP) for processing every point independently. Then, the high-level features of all points are aggregated by a symmetric function (e.g., max, mean, sum etc.), which ensures invariance to the permutation of points. Later, many approaches followed this idea when using deep learning with point clouds.~As in \cite{zhang2019review}, which surveyed many of the recent works, these works can be classified into three categories based on their contributions: (i) contributions related to 3D CNNs; (ii) contributions regarding the network architecture; and (iii) contributions related to neighbor point feature aggregation.

\noindent
\textbf{i. 3D CNN-based methods:} These methods do not process point cloud directly, but instead, they
first convert unordered set of points into a more structured voxel grid, and then perform 3D convolution to obtain the prediction result. 
Although VoxNet\cite{maturana2015voxnet} is not a semantic segmentation method, it is one of the early milestones using deep learning for 3D vision. PointGrid~\cite{le2018pointgrid} is a mixed model integrating points and grids. It employs a 3D CNN to learn the grid cells with fixed points, and to obtain the details of local geometry. 
Different from other methods, Cylinder3D~\cite{zhu2021cylindrical} 
converts point coordinates from Cartesian to cylinder coordinate system, and uses a cylinder 3D convolution to make the prediction.

\begin{figure*}[t] 
\centering 
\includegraphics[width=\textwidth]{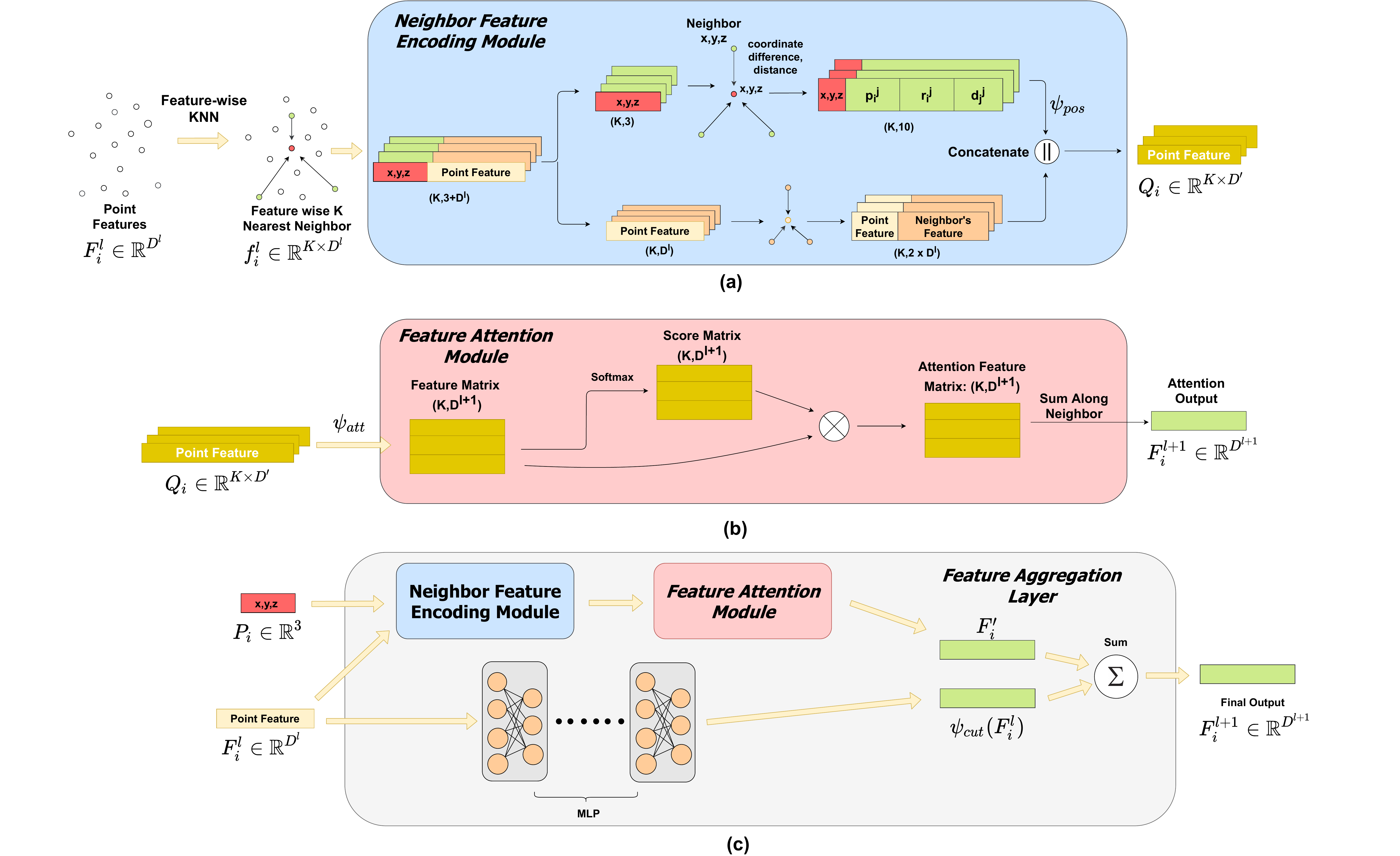} 
\caption{(a) shows the flow of Neighbor Feature Encoding Module. The neighbor's coordinate and feature is processed separately and concatenated in the end. (b) shows the flow of Feature Attention Module, based on the neighbor's feature, each neighbor point has different effect on the query point's feature. (c) shows the whole feature aggregation module. Besides the Neighbor Feature Encoding Module and Feature Attention Module, a shortcut branch is performed to avoid overfitting.} 
\label{feature_agg}
\end{figure*}

\noindent
\textbf{ii. Methods presenting different network architectures:}
In a deep learning-based 3D point cloud analysis task, the receptive field plays a crucial role. If the receptive field is too large, the outcome may be affected by too much of invalid information, whereas if the receptive field is too small, global information may not be effectively captured. PointNet++\cite{qi2017pointnet++}, the successor of PointNet, proposed to exploit the local information with the use of multiple hierarchies with different receptive field sizes. In each hierarchy level, Farthest Point Sampling (FPS) method is performed to downsample the points. In this way, a point's features in different receptive fields could be used to make the final prediction. For each point in different hierarchy levels, maxpooling is performed to gather features of neighboring points. 3DMAX-Net~\cite{ma20183dmax} is another method adopting the idea of multi-scale. It consists of two core parts, namely Multi-Scale Contextual Feature Learning Block (MS-FLB) and Local and Global Feature Aggregation Block (LGAB). It first fuses the features learned at multiple scales, and then aggregates the local and global features to improve the final accuracy of segmentation. DGCNN~\cite{wang2019dynamic} proposed a point cloud learning method based on a dynamic neighborhood. For each point, instead of defining a single neighborhood by using point coordinates in 3D metric space, DGCNN generates a new neighborhood in the feature space of the output of every layer. In this way, neighbors of a point change as the point propagates through the network layers.

\noindent
\textbf{iii. Contributions related to neighbor feature aggregation:} PointNet~\cite{qi2017pointnet} has shown that maxpooling layer can effectively make the model invariant to point permutation. Compared with other following work, PointNet has a relatively simple network architecture, which only contains an MLP. An Annularly Convolutional Neural Network (A-CNN) was proposed in \cite{komarichev2019cnn}, which is a hierarchical model for semantic segmentation of large scenes. A graph attention convolution method was presented in \cite{wang2019graph} to aggregate features from the neighbor points of target points. By the graph attention method, the effect of different neighbors to query point is learned. RandLA-Net~\cite{hu2020randla} uses a feature aggregation method consisting of two parts, namely Local Spatial Encoding (LocSE) and Attentive Pooling. In RandLA-Net, first, neighbor point features and coordinates are processed separately in the LocSE module. Then, the Attentive Pooling aggregates the information of the processed neighborhood. Compared to many other works, RandLA-Net can handle a large number of points at once, and its feature aggregation module is relatively complex. Besides, it is based on PointNet++ structure, which slows down the running speed of the model.~DGCNN\cite{wang2019dynamic} is more efficient, yet its dynamic neighborhoods only make use of the higher dimensional features, and ignore the raw positional information of the points in the later layers. To address the aforementioned problems, our proposed approach encodes the higher dimensional features and position information of a dynamic neighborhood in two separate branches at the beginning, and then aggregates them by attention pooling. We show that this approach can obtain a better feature representation for query points, and achieve the state-of-the-art overall accuracy for semantic segmentation.

\section{Dynamic Point Feature Aggregation}
\label{sec:dynamic-point-feature-attention}

It has been shown that point cloud learning models highly benefit from the point neighborhood for capturing local information \cite{hu2020randla,zhang2019shellnet,zhao2019pointweb}. Yet, the neighborhood information may become an additional hurdle without an effective aggregation strategy. Most of the previous works adopted the K-Nearest Neighborhood (KNN) method, to gather the local information from the neighborhood of a point. However, a point being in a close neighborhood of a query point does not mean that this point will provide important local information. On one hand, certain neighbors might be very important indications of what the local region looks like (corner, plane etc.). On the other hand, some of the close neighbor points might be irrelevant for the local semantic region, e.g. neighbor points belonging to a different object. Thus, treating all the neighbors as equal and performing an unweighted aggregation, such as summation, cannot be the best approach.

We propose the Dynamic Point Feature Aggregation (DPFA) for more attentive neighborhood feature aggregation in point cloud learning. The details of the core module, which is the Feature Aggregation (FA) layer, are presented in Sect.~\ref{ssec:feature-aggregation-module}. Then, two variants of a new network (DPFA-Net), which utilizes FA Layers are presented for semantic segmentation and object classification in Sect.~\ref{ssec:network-architecture}.

\subsection{Feature Aggregation Layer}
\label{ssec:feature-aggregation-module}

It has been shown that capturing local information is important in point cloud learning~\cite{wang2019dynamic,zhao2019pointweb,hu2020randla}. Thus, various methodologies have been presented to define local regions, such as Farthest Point Sampling with ball queries~\cite{qi2017pointnet++} or K-Nearest Neighborhood (KNN)~\cite{wang2019dynamic}. In this work, we obtain the local neighborhood for every point by KNN algorithm. Similar to \cite{wang2019dynamic}, we adopt a dynamic approach to construct the local neighborhood. More specifically, we apply KNN after every layer to find the new neighbors of every point in the current higher-dimensional feature space. In contrast to previous works, our proposed FA Layer also considers raw coordinates of every point to keep reinforcing the positional information throughout the network. Therefore, the proposed FA layer is used to effectively aggregate the position-aware dynamic neighbor features.

Our proposed Feature Aggregation layer is composed of two modules: Neighbor Feature Encoding Module and Feature Attention Module, which are shown in Fig.~\ref{feature_agg}(a) and Fig.~\ref{feature_agg}(b), respectively. In the Neighbor Feature Encoding Module, the positional information from the raw coordinates and higher dimensional features of a neighborhood (obtained by feature-wise KNN as described above) are processed in different branches to retain the raw geometric properties throughout the network. Then, they are concatenated to obtain position encoded-neighborhood features and sent to the Feature Attention Module. Since each neighbor should have different significance (or effect score on the query point), and in order to weigh neighbor points differently, we employ an attention mechanism in the Feature Attention module. This allows different points in the neighborhood selectively contribute to the final features of the query point.

\begin{figure*}[t] 
\centering 
\includegraphics[width=\textwidth]{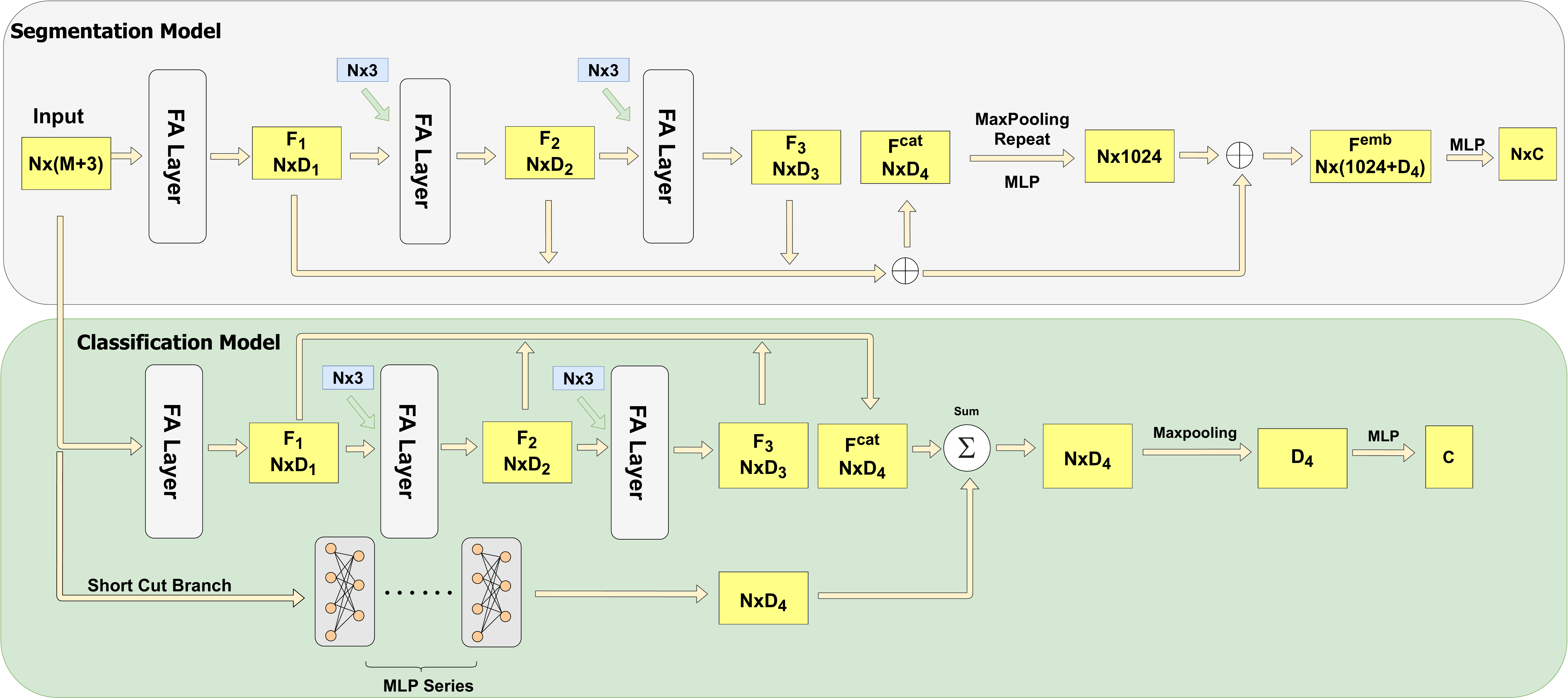} 
\caption{DPFA-Net architectures for semantic segmentation and classification tasks} 
\label{fig:overall-architecture}
\end{figure*}

\subsubsection{Neighbor Feature Encoding Module}

The inputs of this module are the point cloud $\bm{P}\in\mathbb{R}^{N\times3}$ itself and the corresponding three dimensional neighborhood feature graph $\bm{G}\in\mathbb{R}^{N\times K\times D^l}$ of the point cloud, where $N$ is the number of points in the point cloud, $K$ is the neighborhood size and $D^l$ is the feature dimension of layer $l$. Each point in $\bm{G}$ contains its own K-nearest neighborhood information, $\bm{f}_i=[\bm{f}_i^1;\bm{f}_i^2;\cdots;\bm{f}_i^K]^\mathrm{T}$, where $\bm{f}_i^1$ is the features of the point itself (query point). The information from the $j^{th}$ neighbor of point $i$, $\bm{f}_i^j$, is a $D^l$-dimensional vector. For the initial input ($l=1$), $D^1=6$ and $\bm{f}_i^j=[x,y,z,r,g,b]^\mathrm{T}\in\mathbb{R}^{6}$ contains the $x,y,z$ coordinates and RGB colors. For $l>1$, $\bm{f}_i^j\in\mathbb{R}^{D^l}$ is a $D^l$-dimensional vector.
The proposed Neighbor Feature Encoding Module processes the input in two separate branches, namely \textit{Positional Encoding} and \textit{Feature Encoding}, which are described next.

\begin{itemize}[leftmargin=*]
  \item \textbf{Positional Encoding}:  The top branch in Fig.~\ref{feature_agg}(a) shows the flow of positional encoding. 3D coordinates, $\bm{p}^j_i \in\mathbb{R}^{3}$ for $j=\{1,\cdots,K\}$, of the neighborhood of the query point $\bm{p}_i$ is fed as the input. Then, the relative position, $\bm{r}_i^j$, and the distance of each neighbor $j$ to the query, $d_i^j$, are obtained as 
  
\begin{equation}
\bm{r}_i^j=\bm{p}^j_i -\bm{p}^1_i 
\end{equation}
\begin{equation}
d_i^j=\left | \bm{r}_i^j \right |
\end{equation}
  
 where $\bm{p}_i^j$ is the coordinates of the $j^{th}$ neighbor of the $i^{th}$ query point, and $j=1$ refers to the query point itself. The coordinates, relative positions, and relative distances of a query point's neighborhood are concatenated and encoded as follows: 
  
\begin{equation}
\bm{c}_i^j=\psi_{pos}(\bm{p}^j_i \parallel  \bm{p}_i^j \parallel \bm{r}^j_i \parallel d_i^j)
\end{equation}

where ``$\parallel$" represents feature-wise concatenation operation, and the encoder function $\psi_{pos}$ is an MLP, operating on every point. The output of the MLP operation, $\bm{c}_i^j$, is the positional encoding of the $i^{th}$ point's $j^{th}$ neighbor.

\item \textbf{Feature Encoding}
  The bottom branch of Fig.~\ref{feature_agg}(a) shows the flow of feature encoding. $D^l$-dimensional features of the neighborhood of the query point is fed to this branch. Then, the features of the query point is aggregated with every neighbor as follows: 
    \begin{equation}
        \bm{e}_i^j=\phi(\bm{f}_i^1,\bm{f}_i^j)
    \end{equation}
  The aggregation function $\phi$ can be various functions. In our experiments, we have used concatenation ($\parallel$) as the aggregation function. $\bm{e}_i^j$ is the feature of the $j^{th}$ neighbor of the query point $\bm{p}_i$ after feature encoding.

\end{itemize}

After we obtain both the positional encoding $\bm{c}_i$ and the feature encoding $\bm{e}_i$ of a query point's neighborhood, the output of the Neighbor Feature Encoding Module $\bm{Q}_i^j\in D'$ is obtained as:
    \begin{equation}
        \bm{Q}_i^j=\phi(\bm{c}_i^j,\bm{e}_i^j)
    \end{equation}
  
\subsubsection{Feature Attention}
  Fig.~\ref{feature_agg}(b) shows the flow of the Feature Attention Module.
  Encoded features $\bm{Q}\in\mathbb{R}^{N\times K\times D'}$, obtained from the Neighbor Feature Encoding Module, are processed by another MLP, $\psi_{att}:\mathbb{R}^{K\times D'}\rightarrow \mathbb{R}^{K\times D^{l+1}}$, to obtain the feature map $\bm{R}\in\mathbb{R}^{K\times\ D^l}$ of the K-neighborhood of each point, wherein $D^l$ is the feature size of layer $l$. To obtain each neighbor's effect weight for the query point, feature score matrix $\bm{\widehat{R}}$ is obtained by a softmax normalized MLP as follows:
  
    \begin{equation}
        \bm{\widehat{R}}_i=\sigma(\underbrace{\psi_{att}(\bm{Q}_i)}_{\bm{R}_i}), \quad \forall i\in\left \{ 1,2,\cdots,N \right \}
    \end{equation}

  where $\sigma$ is the softmax function and $\psi_{att}$ is the MLP attention encoder. Then, the attentive neighborhood features $\bm{Q'}$ are obtained by the Hadamart product, i.e. element-wise multiplication, of $\bm{R}_i$ and $\bm{\widehat{R}}_i$ as follows:
  
  \begin{equation}
      \bm{Q'}_i=\bm{R}_i \odot \bm{\widehat{R}}_i, \quad \forall i\in\left \{ 1,2,\cdots,N \right \}
  \end{equation}
  
Finally, the neighborhood features are aggregated using summation along the neighbor axis, and the feature attention output $\bm{F'}\in\mathbb{R}^{N\times D^l}$ is obtained as:
  
  \begin{equation}
      \label{eqn:feature-attention-output}
      \bm{F'}_i=\sum_{j=1}^K{\bm{{Q'}}_i^j}, \quad \forall i\in\left \{ 1,2,\cdots,N \right \}.
  \end{equation}

\subsubsection{Shortcut Connection}
Neighborhood coordinate and feature graphs go through the Neighbor Feature Encoding and Attention modules, and then the output $\bm{F'}\in\mathbb{R}^{N\times D^l}$ is obtained as in Eq.~(\ref{eqn:feature-attention-output}). In order to avoid overfitting and help the network retain low-level features, another shortcut branch, which is modeled by another MLP, $\psi_{cut}:\mathbb{R}^{N\times D^{l}}\rightarrow \mathbb{R}^{N\times D^{l+1}}$, is used. The final output of one feature aggregation (FA) layer is obtained by summing the attentive features and shortcut features:
        \begin{equation}
            \bm{F}=\bm{F'}+\psi_{cut}(\bm{P})
        \end{equation}

\noindent Fig.~\ref{feature_agg} (c) shows the complete FA layer.

\begin{table*}[t]
\centering
\caption{Semantic segmentation performance of previous works on the S3DIS dataset\cite{armeni2017joint}. OA stands for overall accuracy.}
\label{tab:bg-segmentation-motivation}
\resizebox{1\textwidth}{!}{%
\begin{tabular}{l|cc|ccccccccccccc}
\toprule[1.0pt]
 & OA (\%) & mIoU (\%) & ceiling & floor & wall & beam & column  & window & door & table & chair & sofa & bookcase & board & clutter \\
\toprule[1.0pt]
PointNet \cite{qi2017pointnet} & 77.85 & 36.31 & 87.01 & 96.68 & 68.41 & 0 & 9.90 & 42.67 & 9.91 & 48.39 &30.35 &2.28 &30.2 &13.72 &32.55\\

PointNet++ \cite{qi2017pointnet++} & 79.51 & 41.14 & 88.01 & 97.43 & 61.28 & 0 & 5.74 & 38.53 & 15.16 &65.06& 64.41& 0.04& 40.32& 19.28& 39.60\\
DGCNN\cite{wang2019dynamic} & 82.62&39.50&91.96&98.73&66.76&0.12&9.07&28.93&3.22&64.61&55.53&9.24&32.04&13.43&39.92 \\

PVCNN\cite{liu2019point} & 85.37&47.44&92.10&98.80&75.13&0&19.24&43.83&21.62&69.28&72.70&4.29&52.83&19.13&47.57\\

 \bottomrule[1.0pt]
\end{tabular}%
}
\end{table*}

\begin{figure*}[t!]
    \centering
    \begin{minipage}{0.42\linewidth}
        \centering
        \includegraphics[width=1.0\linewidth]{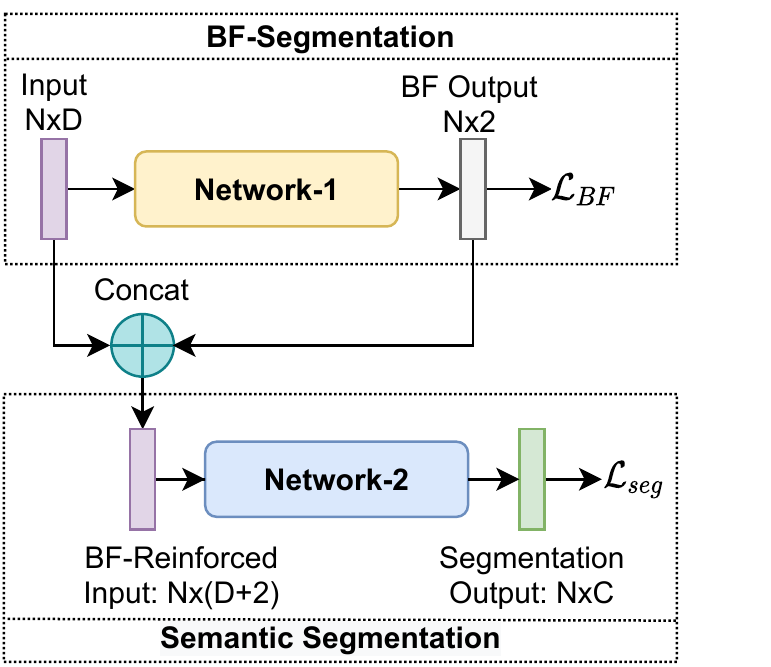}
        (a) Two-Stage BF-Net
    \end{minipage}
    \begin{minipage}{0.47\linewidth}
        \centering
        \hspace*{0.5cm}
        \includegraphics[width=1.0\linewidth]{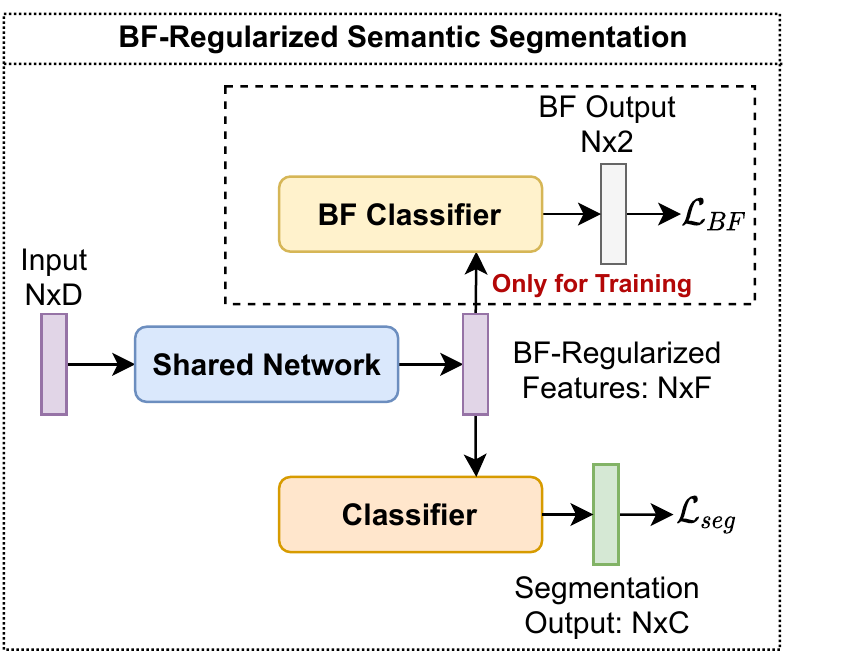}
        \hspace*{-2cm}
        (b) BF-Regularization
    \end{minipage}
    \vspace{1em}
    \caption{The proposed (a) Two-Stage BF-Net (left) and (b) BF Regularization networks. In Two-Stage BF-Net, Network-1 is responsible for binary Background-Foreground segmentation, the output of which is concatenated to the input feature vector of the point cloud. This new BF-Reinforced feature vector is fed into Network-2 for final segmentation. Network-1 and Network-2 are trained separately. In BF Regularization, a single shared network is used for feature extraction. During training, a separate BF classifier is used to regularize the learning in favor of increasing the inter-class variance of feature vectors of background and foreground categories in the feature space.}
    \label{fig:bf-structure}
\end{figure*}

\subsection{DPFA-Net Architecture}
\label{ssec:network-architecture}

In our proposed architecture, we employ dynamic edge convolution \cite{wang2019dynamic}, which was shown to be superior to static neighborhood pooling, to exploit the local information in the point cloud. More specifically, instead of generating point features directly from their embeddings, edge convolution generates edge features that describe the relationships between a point and its neighbors. Moreover, the neighborhood graph of the point cloud is recalculated at every layer based on the Euclidean distance in the hyperspace of the input features. Therefore, the neighborhood in the original edge convolution is fully dynamic. Moreover, we argue that the raw point coordinates retain the relative positional information for a given point or neighborhood. Thus, in addition to the dynamic features, we also keep the original input coordinates, $P\in\mathbb{R}^{N\times3}$, in every layer, and concatenate them with the dynamic features.

Although the main motivation is the semantic segmentation of the point clouds, the proposed model is also capable of learning information about the entirety of the objects. If the model performs well on the semantic segmentation task, then it should be able to provide good results on the classification task as well. Hence, we propose a feature aggregation-based object classification network for point clouds in addition to the proposed segmentation network. 

The overall architecture of our proposed DPFA-Net for the semantic segmentation and classification tasks are illustrated in Fig.~\ref{fig:overall-architecture}. The input for both networks is a point cloud $\bm{P}\in\mathbb{R}^{N\times(3+M)}$, where $N$ is the number of points, 3 is for the 3D coordinate vector $[x,y,z]$, and $M$ is for additional input features, such as point color (RGB), normalized coordinates, normal vectors etc.

In the segmentation network, the input point cloud and the intermediate features are processed by three consecutive Feature Aggregation (FA) layers, which are described in Sect.~\ref{ssec:feature-aggregation-module}.~The outputs of these three FA layers, $\bm{F}^1\in\mathbb{R}^{N\times D^1}$, $\bm{F}^2\in\mathbb{R}^{N\times D^2}$ and $\bm{F}^3\in\mathbb{R}^{N\times D^3}$, are concatenated as $\bm{F}^{cat}=(\bm{F}^1 \parallel \bm{F}^2 \parallel \bm{F}^3)\in\mathbb{R}^{N\times (D^1+D^2+D^3)}$. Then, we aggregate all points by the symmetric \textit{maxpool} operation to get a permutation invariant embedding of the entire point cloud, $\bm{E}\in\mathbb{R}^{(D^1+D^2+D^3)}$, and concatenate this global embedding with every point to get globally-aware point embeddings: $\bm{F}_i^{emb}=(\bm{F}_i^{cat} \parallel \bm\psi_{emb}({E}))\ \forall i\in\left \{ 1,2,\cdots,N \right \}$, where \bm $\psi_{emb}:\mathbb{R}^{(D^1+D^2+D^3)}\rightarrow\mathbb{R}^{1024}$ is modeled as an MLP and  $\bm{F}_i^{emb}\in\mathbb{R}^{(1024+D^1+D^2+D^3)}$.

The classification network, shown in the lower part of Fig.~\ref{fig:overall-architecture}, is similar to the segmentation network. Three consecutive FA layers process the input cloud and the intermediate features, and these intermediate features are concatenated together. Different from segmentation, a shortcut branch is used in the classification network to avoid overfitting. The output of the shortcut branch, which is modelled by an MLP $\psi_{ccut}$, is added to the concatenated point features. These summed features are pooled along the point dimension to get global point cloud embedding $\bm{E}^C\in\mathbb{R}^{D^{pool}}$, and the final output of the classification network is obtained by a classifier MLP, $\psi_{cls}:\mathbb{R}^{D^{pool}}\rightarrow \mathbb{R}^{C}$.
    \begin{equation}
        C_{out}=\sigma(\psi_{cls}(\bm{E}^C+\psi_{ccut}(\bm{P})))\in\mathbb{R}^C,
    \end{equation}
\noindent where $C$ is the number of classes, and $\sigma$ is the \textit{softmax} operation to get the class probabilities.

\section{Background-Foreground Segmentation}
\label{sec:bg-segmentation}

We have observed a common outcome from the earlier studies that the semantic segmentation models perform consistently well for certain classes. More specifically, all models tend to perform better on semantic labels, which are attributed to more structural, relatively larger and static elements of the scene, such as walls, floors and ceiling, than on the other less common labels with more intra-class variation. Table~\ref{tab:bg-segmentation-motivation} shows the segmentation performance of our experiments performed with four different baseline models on the S3DIS dataset~\cite{armeni2017joint}. This dataset contains point cloud captures of 271 rooms from 6 indoor scenes. In S3DIS, there are 13 different labels belonging to different semantic categories. As seen in Table~\ref{tab:bg-segmentation-motivation}, even though different models have different scores for the same classes, they have one thing in common, which is the fact that all models perform much better on ceiling, floor and wall classes compared to other categories. One reason for this is that these three types of classes have relatively simpler structure being mostly planar. Additionally, they have greatly reduced intra-class variance and are significantly over-represented in the dataset. This fact of consistently high performance of different models on these classes is exploited in this paper to further improve the semantic segmentation model performance.

We define those categories, which are similar in nature to walls, ceiling, floor etc., as the background and the rest of the classes as the foreground. In this paper, we propose two ways of handling the Background-Foreground (BF) information. These methods are referred to as \textit{two-stage BF-Net} and \textit{BF Regularization}, and their structures are shown in Fig.~\ref{fig:bf-structure}.

\subsection{Two-Stage BF-Net}
Two-Stage BF-Net employs two separate segmentation networks.~The first network, referred to as Network-1, performs the background-foreground classification. The one-hot-encoded binary classification output of the Network-1 ($[0,1]$ or $[1,0]$) is concatenated to the input feature vector (point coordinates, normalized coordinates, RGB color, normal vector etc.) of the corresponding point. Then, the resulting vector is fed into the second segmentation network, referred to as Network-2, to get the final segmentation result. The structure of Two-Stage BF-Net is shown in Fig.~\ref{fig:bf-structure}(a). This approach allows using any off-the-shelf segmentation model as Network-1 or Network-2, or for both. As supported by the experimental results presented in Sec.~\ref{ssec:bfnet-experiments}, providing the BF information not only increases the accuracy of the second network, but also accelerates the convergence speed of Network-2 during training.

In this Two-Stage BF-Net, Network-1 needs to be pre-trained for the BF classification task. Network-1 is trained via back-propagation using binary cross entropy as the loss function. Once Network-1 is trained, it is utilized in the training of the Network-2. In this second training, while Network-2 is updated via the back-propagation using the cross entropy loss, Network-1 is not updated.

\begin{table*}[ht]
\centering
\caption{Semantic segmentation results on S3DIS dataset\cite{armeni2017joint} evaluated on Area 5}
\label{s3dis_area5}
\resizebox{1\textwidth}{!}{%
\begin{tabular}{lcc|ccccccccccccc}
\toprule[1.0pt]
 & OA (\%) & mIoU (\%) & ceiling & floor & wall & beam & column  & window & door & table & chair & sofa & bookcase & board & clutter \\
\toprule[1.0pt]
PointNet \cite{qi2017pointnet} & - & 41.09 & 88.80 & 97.33 & 69.80 & 0.05 & 3.92 & 46.26 & 10.76 & 58.93 &52.61 &5.85 &40.28 &26.38 &33.22\\

SegCloud \cite{tchapmi2017segcloud} & - & 48.92 & 90.06 & 96.05 & 69.86 & 0.00 & 18.37 & 38.35 & 23.12 & 70.40 & 75.89 & 40.88& 58.42& 12.96& 41.60\\
PointCNN \cite{li2018pointcnn} & 85.91 &57.26& 92.31 &98.24 &79.41&0.00 &17.60 &22.77 & 62.09& 74.39 & 80.59&31.67 &66.67&62.05&56.74 \\

SPGraph\cite{landrieu2018large} & 86.38 &58.04 &89.35 &96.87&78.12 & 0.00 &42.81&48.93 &61.58 &84.66 &75.41 & 69.84&52.60 &2.10&52.22\\

PCCN\cite{wang2018deep} & - & 58.27 &92.26  &96.20 &75.89 &0.27 &5.98 &69.49  &63.45 & 66.87 &65.63 & 47.28 &68.91 & 59.10&46.22\\

PointWeb \cite{zhao2019pointweb} & 86.97 & \textbf{60.28} &91.95  &98.48  &79.39 &0.00 &21.11 &59.72   &34.81 & 76.33 &88.27 &46.89 &69.30  &64.91 & 52.46\\

DPFA(ours) & \textbf{87.47} & 52.96 & 93.7 & 98.72 & 75.50 & 0.00& 14.36 & 50.14 & 31.76 & 73.68 & 73.36 & 13.68 & 55.50 & 57.05& 51.18\\

DPFA+BF Reg(ours) & \textbf{88.04} & 55.19  &92.95 & 98.55 & 80.23 & 0& 14.72 & 55.84 & 42.79 & 72.29 & 73.50 & 27.26 &55.86& 52.97 & 50.50\\

 \bottomrule[1.0pt]
\end{tabular}%
}
\end{table*}

\begin{table*}[ht]
\centering
\caption{Semantic segmentation results on S3DIS dataset\cite{armeni2017joint} with 6-folds cross validation.}
\label{s3dis_6folds}
\resizebox{1\textwidth}{!}{%
\begin{tabular}{lcc|ccccccccccccc}
\toprule[1.0pt]
 & OA (\%) & mIoU (\%) & ceiling & floor & wall & beam & column  & window & door & table & chair & sofa & bookcase & board & clutter \\
\toprule[1.0pt]
PointNet \cite{qi2017pointnet} & 78.5  & 47.6 & 88.0 & 88.7 & 69.3 & 42.4 & 23.1 & 47.5 & 51.6 &  54.1 &42.0  &9.6 &38.2 &29.4 &35.2\\

PointCNN \cite{li2018pointcnn} & 88.1 &65.4 & 94.8 &97.3 &75.8 &63.3 &51.7 &58.4 & 57.2& 71.6 & 69.1 &39.1 &61.2&52.2&58.6 \\

SPGraph\cite{landrieu2018large} & 85.5  &62.1 &89.9  &95.1 &76.4  &62.8  &47.1 &55.3   &68.4 &73.5 &69.2 & 63.2 &45.9 &8.7 &52.9 \\

PointWeb \cite{zhao2019pointweb} & 87.31 & 66.73  &93.54  &94.21  &80.84 &52.44 &41.33 &64.89   &68.13 & 71.35 &67.05 &50.34 &62.68  &62.20 & 58.49\\

ShellNet \cite{zhang2019shellnet} &87.1&66.8&90.2&93.6&79.9&60.4&44.1&64.9&52.9&71.6&84.7&53.8&64.6&48.6&59.4 \\

DGCNN\cite{wang2019dynamic} &84.1 & 56.1 &-&-&-&-&-&-&-&-&-&-&-&-&-\\

RandLA-Net\cite{hu2020randla} & 88.0 & \textbf{70.0} &93.1 & 96.1 & 80.6 & 62.4 & 48.0 & 64.4 & 69.4 & 69.4 & 76.4 & 60.00 & 64.2 & 65.90 & 60.1 \\

DPFA(ours) & \textbf{89.01} & 61.61  &94.61 & 97.68 & 77.84 & 38.45& 38.28 & 53.34 & 67.66 & 66.60 & 75.23 & 29.48 &49.79& 51.38& 60.64\\

DPFA+BF Reg(ours) & \textbf{89.22} & 61.70  &94.58 & 98.00 & 79.20 & 40.65& 36.55 & 52.16 & 70.76 & 65.89 & 74.69 & 27.68 &49.80& 51.55& 60.64\\

 \bottomrule[1.0pt]
\end{tabular}%
}
\end{table*}

\subsection{BF Regularization}

Although the semantic segmentation benefits from the BF information, Two-Stage BF-Net requires a well-selected Network-1 and separate training processes for Network-1 and Network-2, which may not always be very practical. In order to address this, we have designed the BF-Regularization method, which can be applied to any semantic segmentation model. BF-Regularization addresses the issue of separate training processes in Two-Stage BF-Net, and provides a simpler architecture with a single-step and end-to-end training. In BF regularization, instead of employing a separate network for classifying background and foreground points, a single backbone network is shared by BF segmentation and semantic segmentation. Then, by formulating the BF learning as the regularizer during training, the model is promoted to inherently increase the inter-class variance between the background and foreground classes in the latent space. In our experiments, we apply this method to our proposed DPFA-based semantic segmentation network and it improves our model's accuracy as will be shown in Sec.~\ref{sec:experimental-results}.

BF Regularization structure is shown in Fig.~\ref{fig:bf-structure}(b). Instead of training two separate networks, BF Regularization approach integrates a BF Classifier as an auxiliary branch to the network, which regularizes the learning in favor of increasing the inter-class variance of feature vectors of background and foreground categories in the feature space. Both BF classification loss $L_{bf}$ and segmentation loss $L_{seg}$ are categorical cross-entropy losses, and $L_{bf}$ has only two categories (binary cross-entropy). The final loss $L$ is calculated as follows: 
    \begin{equation}
        L=(1-\lambda)\cdot L_{seg}+\lambda\cdot L_{bf}
    \end{equation}
\noindent where $\lambda$ is a scalar adjusting the importance of BF regularization.

\section{Experimental Results}
\label{sec:experimental-results}
We have performed the semantic segmentation experiments on the Stanford large-scale 3D Indoor Spaces Dataset (S3DIS)~\cite{armeni2017joint}, which contains 3D point clouds of 271 rooms from 6 different areas. Each point in a point cloud is annotated with one of the following 13 semantic labels representing different categories: \{chair, table, sofa, bookcase, board, window, door, column, beam, floor, wall, ceiling and clutter\}. To prepare the training and testing data, we split each room into blocks of $1m \times 1m \times Zm $, where $Z$ is the height of the room, and 4096 points are sampled from each block. Each point in a block has the following features: the 3D coordinates ($x_r$,$y_r$,$z_r$), RGB color ($r$,$g$,$b$) and the normalized (according to the block origin) coordinates ($x_b$,$y_b$,$z_b$). Thus, the input point cloud for each block becomes $\bm{P_B}\in\mathbb{R}^{4096\times9}$.

We first present and discuss the results obtained by our proposed DPFA-Net with and without the BF regularization on the S3DIS dataset. We also evaluate the effectiveness of our two proposed Background-Foreground (BF) exploitation approaches, described in Section \ref{sec:bg-segmentation}. Then, we present the results of the object classification network, followed by the results for object part segmentation, which is another common type of segmentation application.  Finally, we provide results comparing different approaches in terms of computational efficiency.  

\subsection{Implementation Details}
All our experiments have been performed on GeForce GTX 1080 Ti GPUs. For the semantic segmentation network, we trained the network with mini-batches of size 5, using ADAM optimization~\cite{kingma2014adam}, for 100 epochs. The learning rate is set to 0.001 with 70\% staircase decay every 20 epochs. In FA Layers, we used 20 as the neighborhood size. $D_{1}$,$D_{2}$ and $D_{3}$ are set to 64 for semantic segmentation, and are set to 64, 128 and 256, respectively, for classification.

\subsection{Dynamic Point Feature Aggregation}

\begin{figure*}[t!] 
\centering 
\includegraphics[scale=0.6]{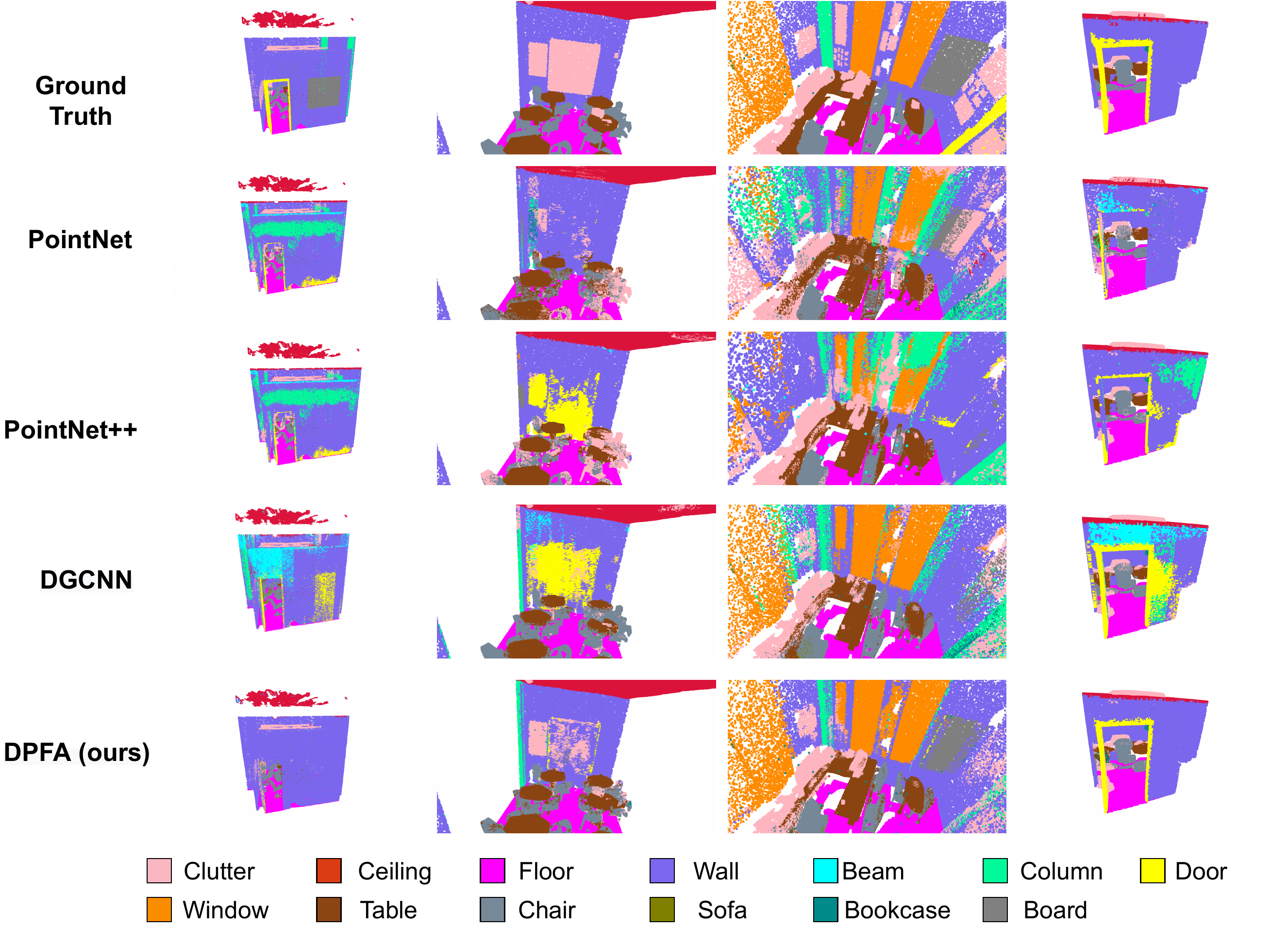} 
\caption{Qualitative examples showing the semantic segmentation results obtained with different approaches. The columns show the segmentation results of four different rooms from the S3DIS dataset.} 
\label{seg_out}
\end{figure*}

Similar to the previous works, we first separated Area-5 of S3DIS for testing, and trained our model on the remaining areas, which is the standard split suggested by the authors of the dataset~\cite{armeni20163d}. In addition to the standard split, we have performed 6-fold cross validation, wherein we leave one area out during the training, test on the left-out area, and then take the average of the results. It should be noted that Area 2 has two auditoriums, whereas there are no auditoriums in the rest of the areas. Hence, to balance the training and testing sets, when testing on Area 2, we have moved one auditorium from the testing set (Area 2) to the training set. The experimental results with the Area-5 testing and 6-fold cross validation are presented in Tables~\ref{s3dis_area5} and \ref{s3dis_6folds}, respectively, for our method as well as different baseline methods. It can be seen from the tables that our method outperforms all the other methods, and achieves the state-of-the-art performance on the Overall Accuracy (OA) metric both with and without BF regularization. \mbox{$O\!A=A/N$}, where $A$ is the number of correctly classified points, and $N$ is the total number of points. The BF regularization further improves the performance. Even though RandLA-Net has a better overall mIoU score for this task, our model is approximately three times faster than RandLA-Net. More specifically, our proposed DPFA takes 28.88 ms and 29.58 ms, without and with BF-Regularization respectively, to process a single input from the S3DIS dataset, while the RandLA-Net takes 88.25 ms to process the same data. A discussion of the run-time performance is provided in Section~\ref{ssec:performance-analysis}. Example qualitative results obtained with different methods are presented in Fig.~\ref{seg_out} to illustrate the effectiveness of our method.

\subsection{Background-Foreground Segmentation}
\label{ssec:bfnet-experiments}
In this subsection, we will present the results of the experiments performed with our proposed BF exploitation approaches, and provide a discussion.

\subsubsection{Two-Stage BF-Net Experiments}

In the Two-Stage BF-Net, Network-1 plays an auxiliary role to improve the second network's performance. Thus, a network with faster speed and reasonable accuracy is preferable to be the first network. In this work, we first experimented with four existing semantic segmentation models, namely PointNet~\cite{qi2017pointnet},  PointNet++~\cite{qi2017pointnet++}, DGCNN~\cite{wang2019dynamic} and PVCNN~\cite{liu2019point}, as Network-1, and evaluated their performance on background-foreground segmentation. The results are displayed in  Table~\ref{tab:bf_seg}. As can be seen, 
PVCNN and PointNet achieve the best and the second best performance in the binary segmentation task, respectively. A possible explanation for PointNet performing better than PointNet++ and DGCNN is that for BF segmentation, global information is more important than local information, and local details are not needed as much. Compared to PointNet++ and DGCNN, PVCNN and PointNet focus more on global features. Based on these observations, we have used both PVCNN and PointNet as Network-1 in our experiments. We have experimented with PointNet, in addition to PVCNN, as the BF segmentation network, since it does not use a graph network or hierarchical structure, and its speed is faster than the other networks.

\begin{table}[hbt!]
    \centering
    \caption{Background/Foreground segmentation performance of four existing models}
    \label{tab:bf_seg}
\label{tab:s3dis}
    \begin{tabular}{|r|c|}
        \hline
         &OA(\%)\\
         \hline
         PVCNN & 92.12\\
         \hline
         PointNet & 90.15\\
         \hline
         PointNet++ & 89.70\\
         \hline
         DGCNN & 86.62\\
         \hline
    \end{tabular}
\end{table}

\begin{table}[hbt!]
    \centering
    \caption{Two-Stage BF-Net experiments with four existing models. Overall accuracy (OA) and mIoU of semantic segmentation results on Area-5 of the S3DIS dataset\cite{armeni2017joint} are presented.}
    \label{tab:overall}
\label{tab:s3dis}
    \begin{tabular}{|r|c|c|}
        \hline
         &OA(\%)&mIoU(\%)\\
         \hline
         PointNet\cite{qi2017pointnet}&77.85&36.31\\
         \hline
         BF-Net (PointNet+PointNet) &\textbf{79.04}& \textbf{38.69}\\
         \hline
         \hline
         PointNet++\cite{qi2017pointnet++}&79.51&41.14\\
         \hline
         BF-Net (PointNet+PointNet++)& 82.46& 45.56\\
         \hline
         BF-Net (PVCNN+PointNet++)& \textbf{85.20}&\textbf{47.99}\\
         \hline
         \hline
         PVCNN\cite{liu2019point}&85.37&47.44\\
         \hline
         BF-Net (PointNet+PVCNN)&84.3&46.91\\
         \hline
         BF-Net (PVCNN+PVCNN)&\textbf{85.47}&\textbf{48.42}\\
         \hline
         \hline
         DGCNN\cite{wang2019dynamic}&82.62&39.5\\
         \hline
         BF-Net (PointNet+DGCNN)&83.89&42.69\\
         \hline
         BF-Net (PVCNN+DGCNN)& \textbf{87.35}& \textbf{51.24}\\
         \hline
    \end{tabular}
\end{table}

As for the second network (Network-2), we have experimented with four different segmentation models (PointNet, PointNet++, DGCNN and PVCNN) to show that this two-stage BF-Net approach can be applied to different models.
The overall accuracy and mIoU of different network configurations are shown in Table~\ref{tab:overall}. 
For the proposed BF-Net, the network names in parentheses correspond to the first and second networks, respectively, of the two-stage approach. In other words,  the notation is `BF-Net (Network-1 + Network-2)'. As can be seen, the proposed BF-Net improves the performance of all four of the original networks both in terms of overall accuracy and mIoU. For instance, the overall accuracy of original PointNet++ is increased from 79.51\% to 82.46\% and 85.20\% when PointNet and PVCNN are used as the first network, respectively. Similarly, the overall accuracy of original DGCNN is increased from 82.62\% to 83.39\% and 87.35\% when PointNet and PVCNN are used as the first network, respectively. BF-Net, employing PVCNN as the first network and DGCNN as the second network, provides the best performance with 87.35\% overall accuracy and 51.24 mIoU. 

We also performed experiments by using PointNet or PVCNN as the first network, and our proposed DPFA-Net as the second network. Using this two-stage approach did not result in further performance increase for the DPFA-Net. Our observation is that, in general, BF segmentation performance of Network-1 should be much higher (about 6 to 7\%) than the overall semantic segmentation accuracy of Network-2, for Network-1 to improve the performance of Network-2. This can be seen by analyzing Tables \ref{tab:bf_seg} and \ref{tab:overall}. For instance, while (PVCNN+PVCNN) performs better than PVCNN, (PointNet+PVCNN) does not. As for our proposed DPFA-Net, its overall semantic segmentation accuracy is already high at 89.01\%, and the BF segmentation performances of the methods presented in Table \ref{tab:bf_seg} are not significantly higher to further boost this number. Thus, for this reason and also due to other advantages of BF Regularizer, we obtained the results in Tables \ref{s3dis_area5} and \ref{s3dis_6folds} by using the BF regularizer approach, which is described in the next section.

\subsubsection{BF Regularization Experiments}
Although the two-Stage BF-Net provides performance improvement on most models, as demonstrated in Table \ref{tab:overall}, it also has some disadvantages. First, the training process is cumbersome, since it involves two separate training steps. Network-1 needs to be pre-trained  for BF-classification task before training Network-2. Second, Network-1 needs to be chosen carefully. As seen in Table~\ref{tab:overall}, if the BF segmentation performance of Network-1 is not sufficiently higher than the segmentation performance of Network-2, it may instead result in some performance drop. Considering these, we developed the BF Regularization approach, which addresses both these issues, and offers an end-to-end semantic segmentation model training.

We have used the BF-Regularization technique together with the DPFA-Net in our experiments. As shown in Fig.~\ref{fig:bf-structure}(b), the BF-Regularization module is plugged in the original DPFA-Net. The experiments are performed the on S3DIS dataset, and the results are shown in  Tables~\ref{s3dis_area5} and \ref{s3dis_6folds}. It can be seen from the tables that BF-Regularized DPFA-Net is performing better than the vanilla DPFA-Net on both the Area-5 test and the 6-Fold (Area) cross validation.


\subsection{Employing DPFA-Net On Other Tasks}

Our proposed DPFA-Net is mainly designed for semantic segmentation applications. However, due its strength in capturing local information in the input data, it can be applied to other tasks, such as 3D object classification and object part segmentation. In this section, we provide additional experimental results on object classification and object part segmentation tasks.

\begin{table*}[tb!]
    \caption{Object Part Segmentation Results on ShapeNet Parts Dataset}
    \label{tab:partsegmentation-table}
    \scriptsize
    \resizebox{1\textwidth}{!}{
    \begin{tabular}{@{}l|c|cccccccccccccccc@{}}
        \toprule
                                                 & mean          & air           & bag            & cap           & car           & chair         & ear           & guitar        & knife         & lamp          & laptop        & motor         & mug           & pistol        & rocket        & skate         & table         \\
                                                 &               & plane         &                &               &               &               & phone         &               &               &               &               &               &               &               &               & board         &               \\ \midrule
        \# shapes                                &               & 2690          & 76             & 55            & 898           & 3758          & 69            & 787           & 392           & 1547          & 451           & 202           & 184           & 283           & 66            & 152           & 5271          \\ \midrule
        PointNet~\cite{qi2017pointnet}           & 83.7          & 83.4          & 78.7           & 82.5          & 74.9          & 89.6          & 73.0          & 91.5          & 85.9          & 80.8          & 95.3          & 65.2          & 93            & 81.2          & 57.9          & 72.8          & 80.6          \\
        PointNet++~\cite{qi2017pointnet++}       & 85.1          & 82.4          & 79.0           & 87.7          & 77.3          & 90.8 & 71.8          & 91.0          & 85.9          & 83.7          & 95.3          & 71.6          & 94.1          & 81.3          & 58.7          & 76.4          & 82.6          \\
        Kd-Net~\cite{klokov2017escape}           & 82.3          & 80.1          & 74.6           & 74.3          & 70.3          & 88.6          & 73.5          & 90.2          & 87.2          & 81.0          & 94.9          & 57.4          & 86.7          & 78.1          & 51.8          & 69.9          & 80.3          \\
        LocalFeatureNet~\cite{shen2017neighbors} & 84.3          & \textbf{86.1} & 73.0           & 54.9          & 77.4          & 88.8          & 55.0          & 90.6          & 86.5          & 75.2          & 96.1          & 57.3          & 91.7          & 83.1          & 53.9          & 72.5          & \textbf{83.8}          \\
        PCNN~\cite{atzmon2018point}              & 85.1          & 82.4          & 80.1           & 85.5          & 79.5          & 90.8          & 73.2          & 91.3          & 86.0          & 85.0          & 95.7          & 73.2          & 94.8          & 83.3          & 51.0          & 75.0          & 81.8          \\
        DGCNN~\cite{wang2019dynamic}             & 85.2          & 84.0          & 83.4           & 86.7          & 77.8          & 90.6          & 74.7          & 91.2          & 87.5          & 82.8          & 95.7          & 66.3          & 94.9          & 81.1          & 63.5          & 74.5          & 82.6          \\ 
        PointCNN~\cite{li2018pointcnn}           & \textbf{86.1}          & 84.1          & \textbf{86.45} & 86.0          & \textbf{80.8} & 90.6          & \textbf{79.7}          & \textbf{92.3} & \textbf{88.4} & \textbf{85.3} & \textbf{96.1} & \textbf{77.2} & \textbf{95.3} & \textbf{84.2} & \textbf{64.2} & \textbf{80.0} & 83.0          \\ \midrule
        
        DPFA-Net (Ours) & 85.5 & 84.5 & 81.1 & \textbf{88.2} & 79.0 & \textbf{90.9} & 69.2 & 91.6 & 87.2 & 83.8 & 95.8 & 70.4 & 92.8 & 82.7 & 63.0 & 77.5 & 81.9 \\ \bottomrule
    \end{tabular}
    }
\end{table*}
\begin{figure*}[t]
    \centering

    \begin{minipage}{1.0\linewidth}
        \centering
        \raisebox{-0.5\height}{\includegraphics[width=0.11\linewidth]{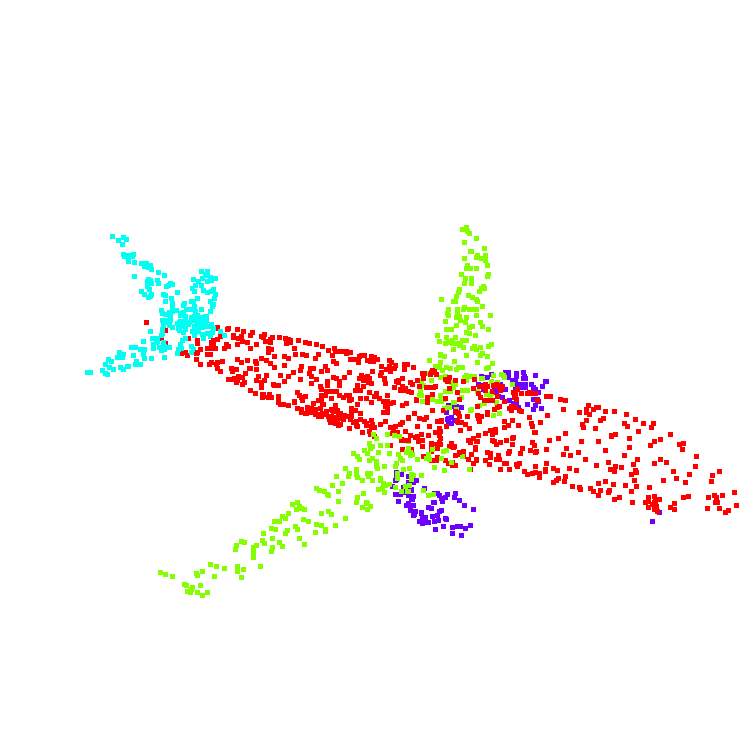}}
        \raisebox{-0.5\height}{\includegraphics[width=0.11\linewidth]{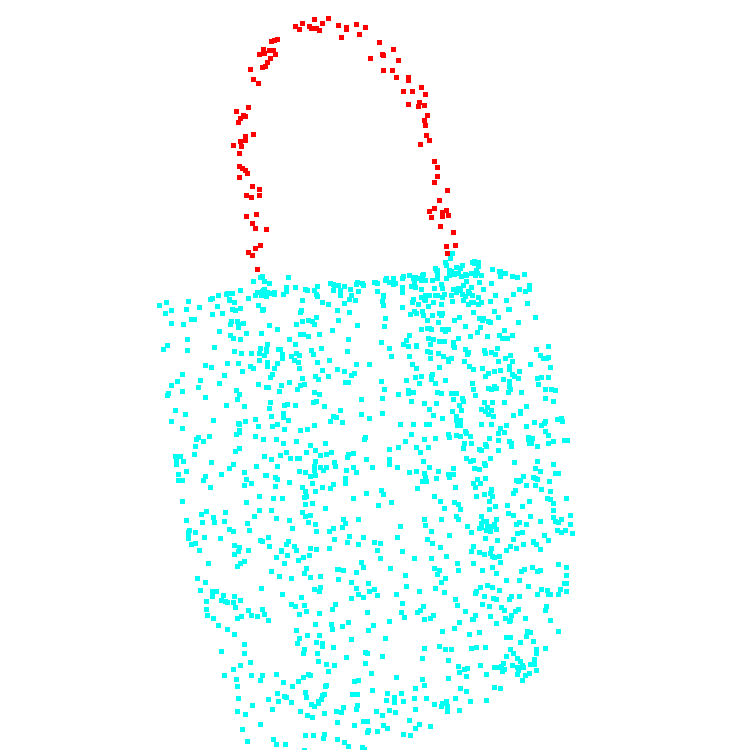}}
        \raisebox{-0.5\height}{\includegraphics[width=0.11\linewidth]{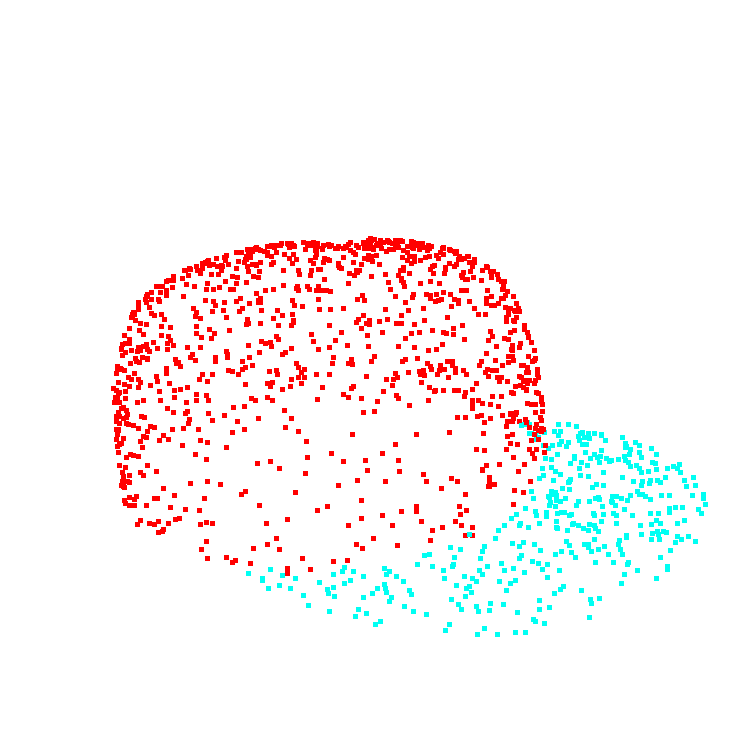}}
        \raisebox{-0.5\height}{\includegraphics[width=0.11\linewidth]{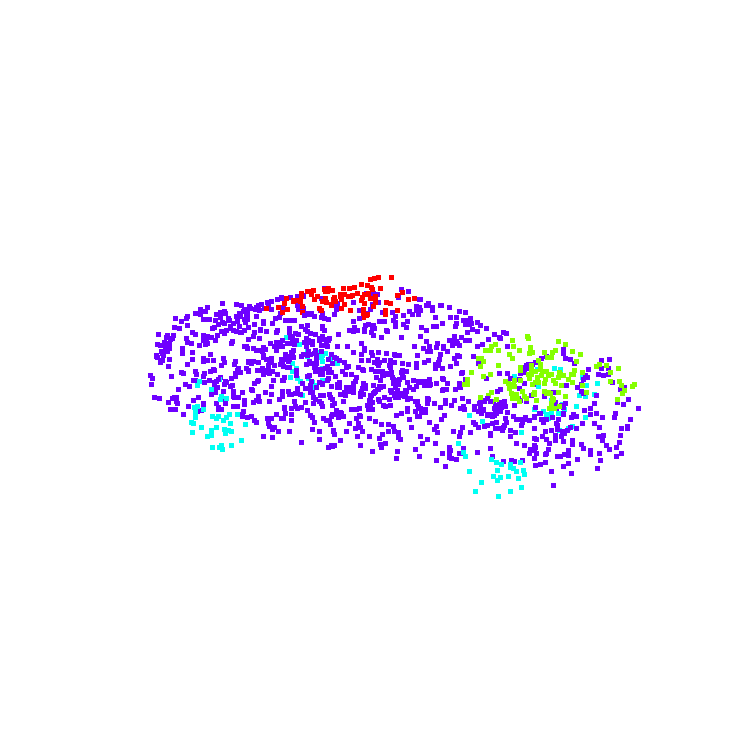}}
        \raisebox{-0.5\height}{\includegraphics[width=0.11\linewidth]{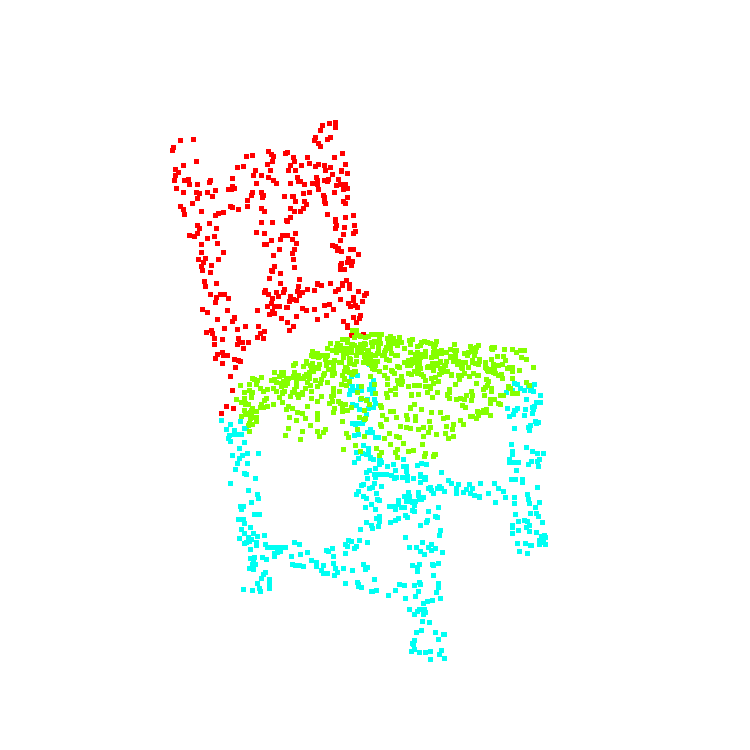}}
        \raisebox{-0.5\height}{\includegraphics[width=0.11\linewidth]{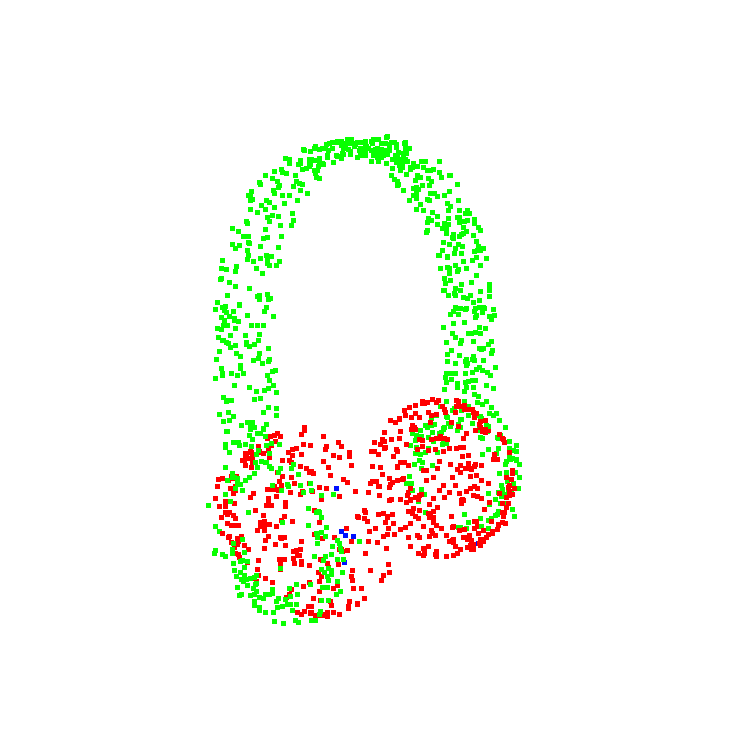}}
        \raisebox{-0.5\height}{\includegraphics[width=0.11\linewidth]{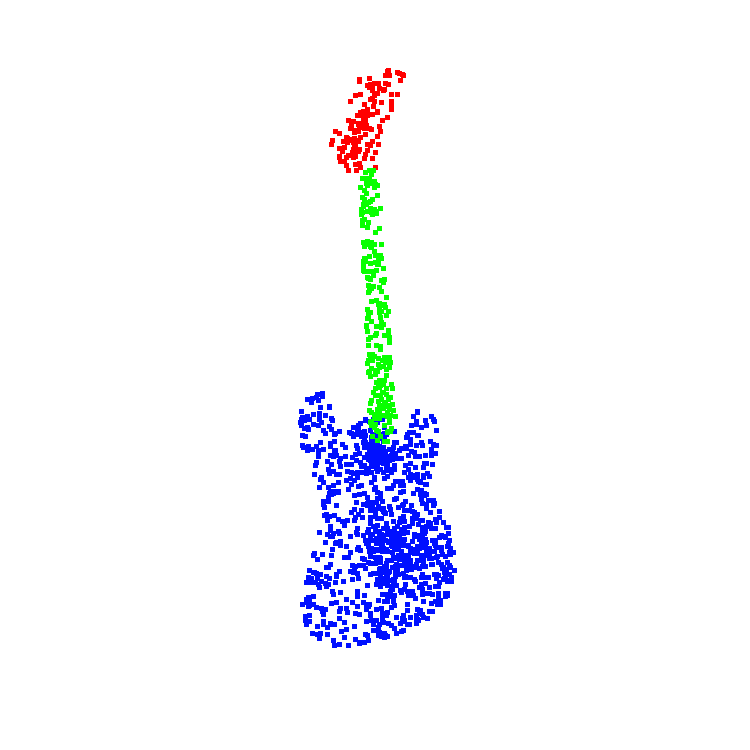}}
        \raisebox{-0.5\height}{\includegraphics[width=0.11\linewidth]{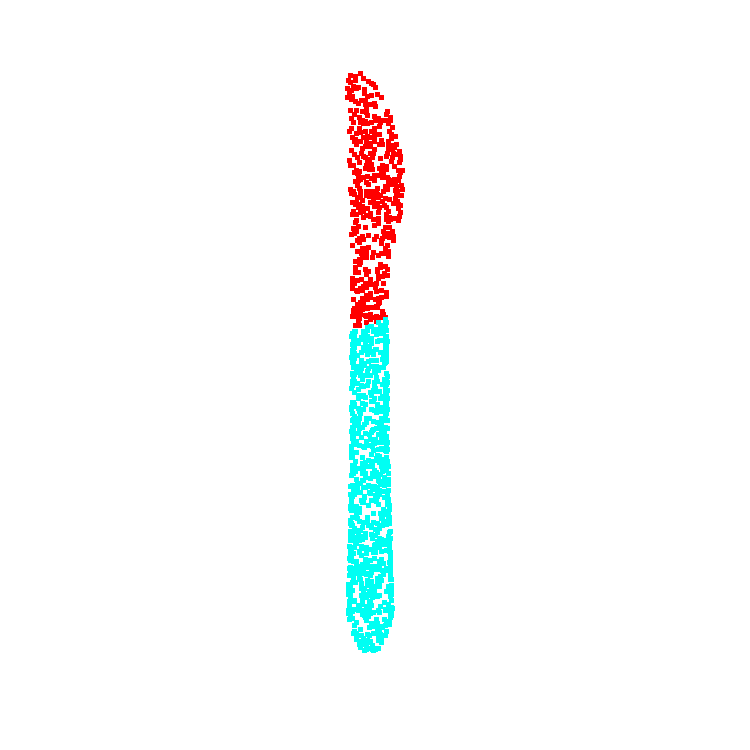}}
      \end{minipage}
      \begin{minipage}{0.11\linewidth}\centering Airplane\end{minipage}
      \begin{minipage}{0.11\linewidth}\centering Bag\end{minipage}
      \begin{minipage}{0.11\linewidth}\centering Cap\end{minipage}
      \begin{minipage}{0.11\linewidth}\centering Car\end{minipage}
        \begin{minipage}{0.11\linewidth}\centering Chair\end{minipage}
        \begin{minipage}{0.11\linewidth}\centering Earphone\end{minipage}
        \begin{minipage}{0.11\linewidth}\centering Guitar\end{minipage}
        \begin{minipage}{0.11\linewidth}\centering Knife\end{minipage}

    \begin{minipage}{1.0\linewidth}
        \centering
        \raisebox{-0.5\height}{\includegraphics[width=0.11\linewidth]{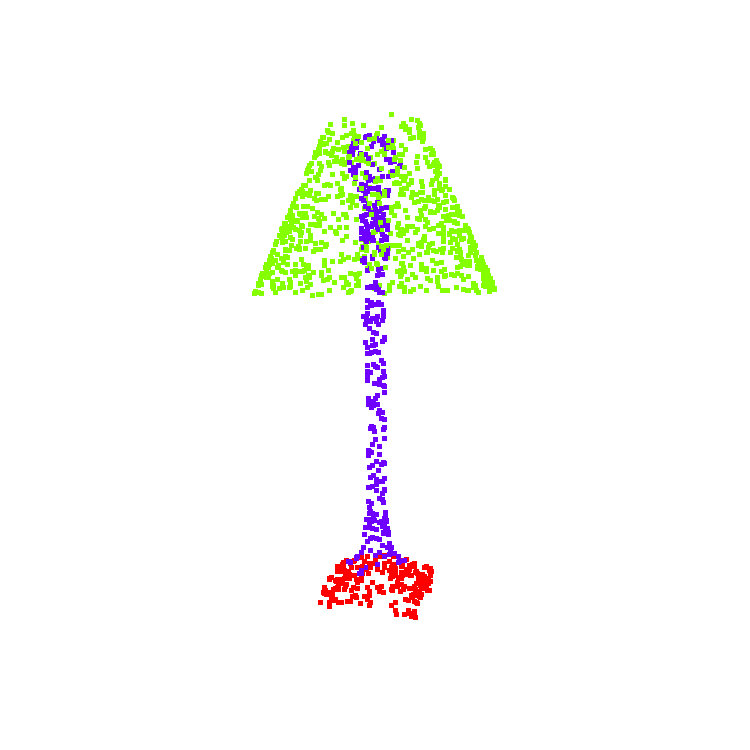}}
        \raisebox{-0.5\height}{\includegraphics[width=0.11\linewidth]{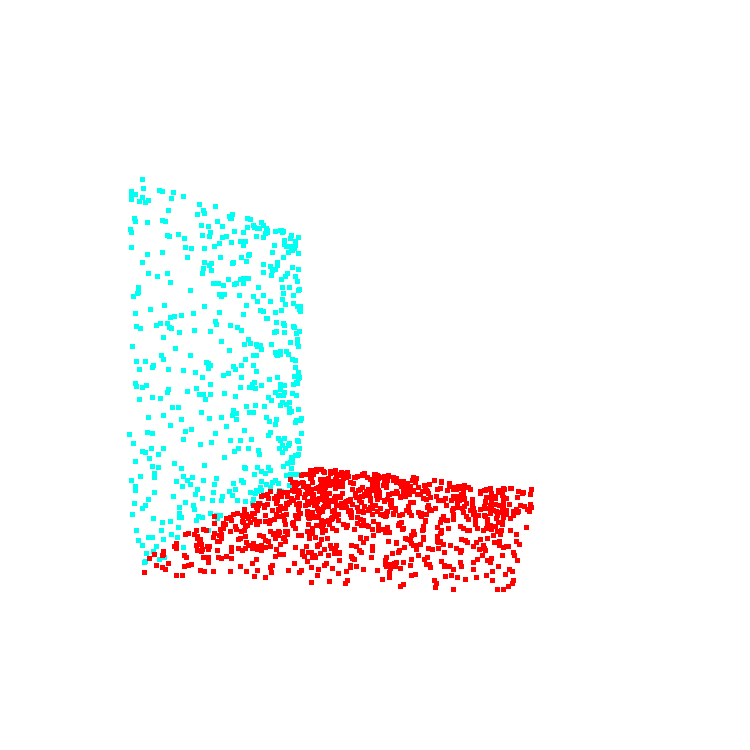}}
        \raisebox{-0.5\height}{\includegraphics[width=0.11\linewidth]{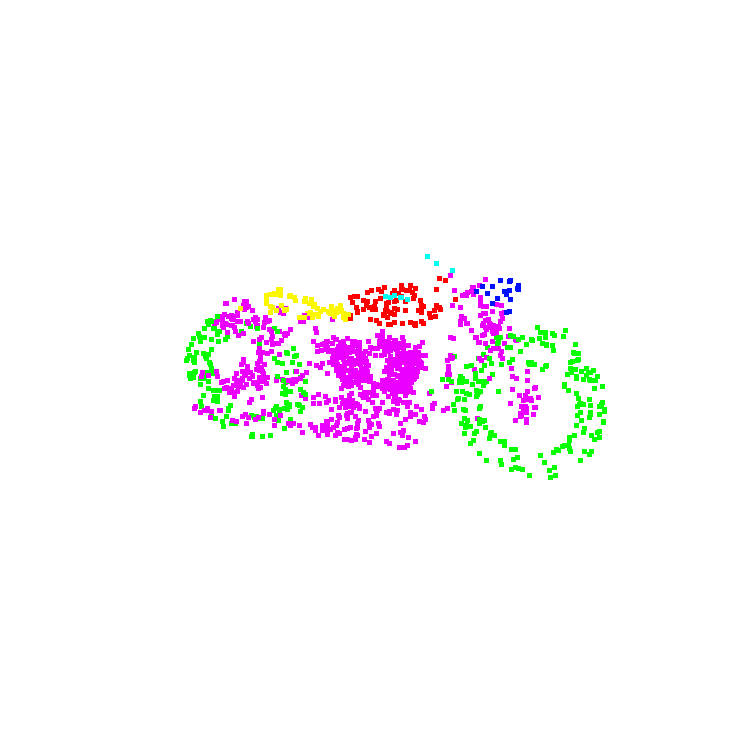}}
        \raisebox{-0.5\height}{\includegraphics[width=0.11\linewidth]{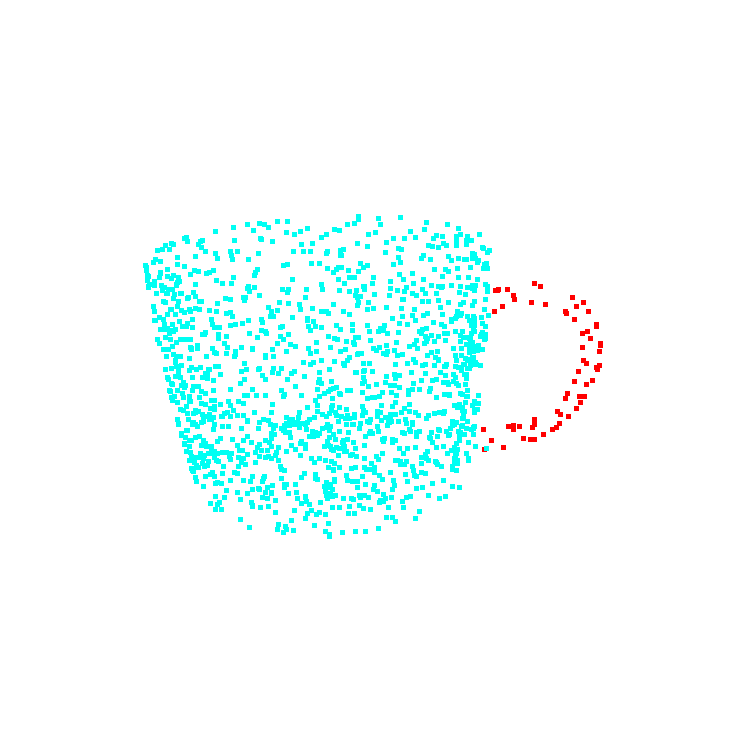}}
        \raisebox{-0.5\height}{\includegraphics[width=0.11\linewidth]{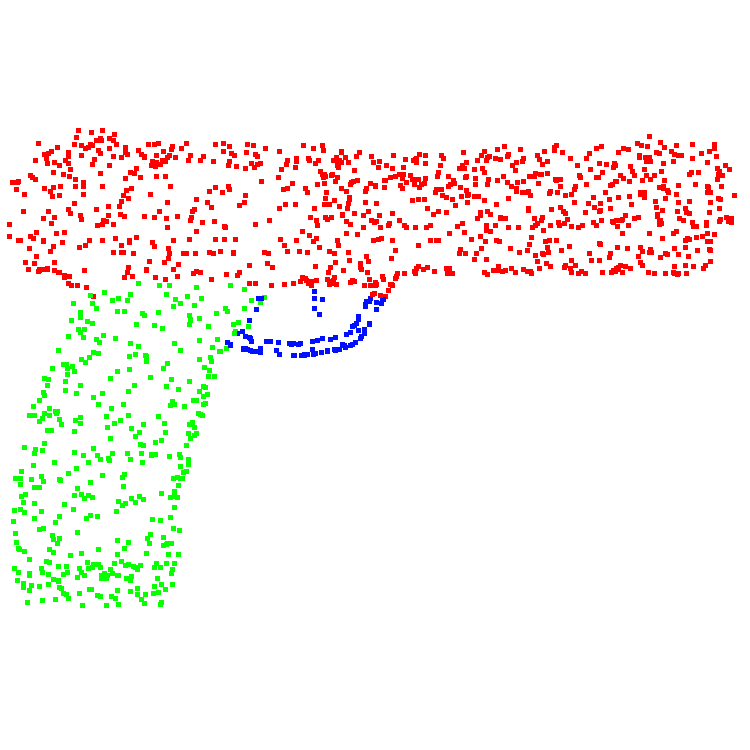}}
        \raisebox{-0.5\height}{\includegraphics[width=0.11\linewidth]{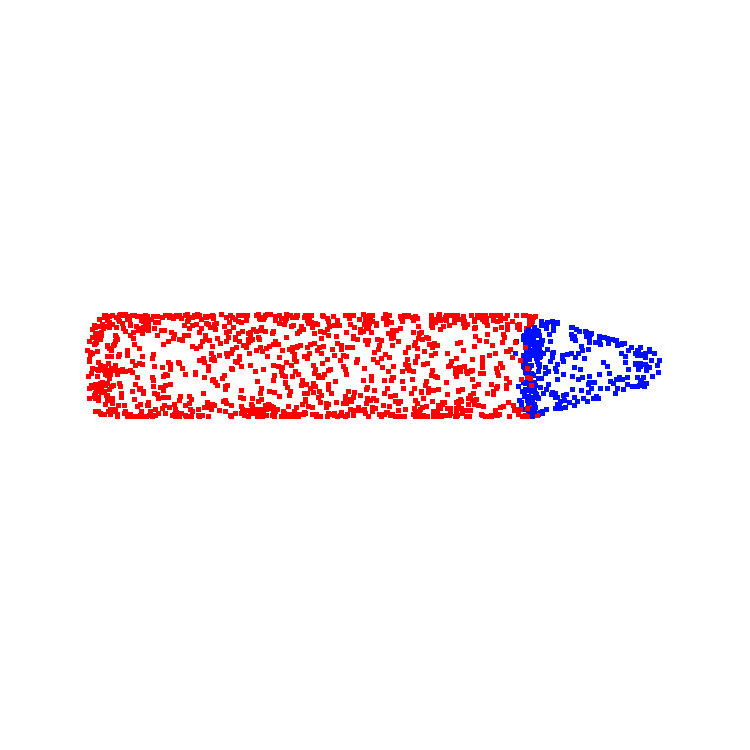}}
        \raisebox{-0.5\height}{\includegraphics[width=0.11\linewidth]{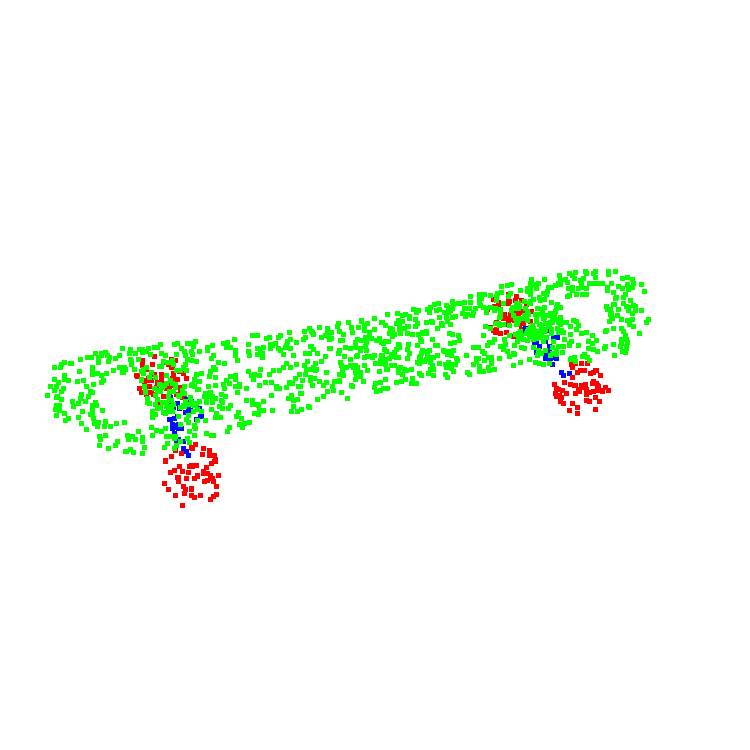}}
        \raisebox{-0.5\height}{\includegraphics[width=0.11\linewidth]{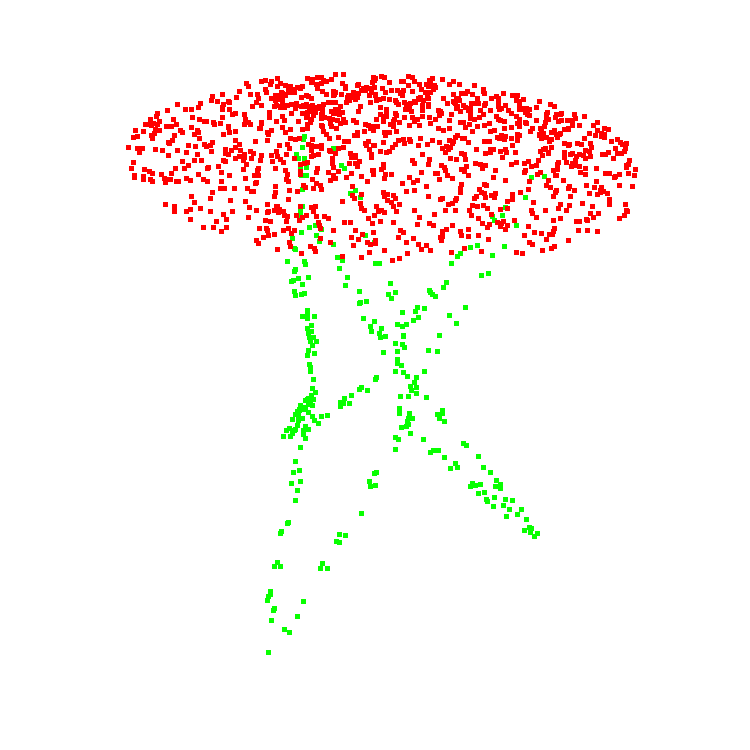}}
        \end{minipage}
        \begin{minipage}{0.11\linewidth}\centering Lamp\end{minipage}
        \begin{minipage}{0.11\linewidth}\centering Laptop\end{minipage}
        \begin{minipage}{0.11\linewidth}\centering Motorbike\end{minipage}
        \begin{minipage}{0.11\linewidth}\centering Mug\end{minipage}
        \begin{minipage}{0.11\linewidth}\centering Pistol\end{minipage}
        \begin{minipage}{0.11\linewidth}\centering Rocket\end{minipage}
        \begin{minipage}{0.11\linewidth}\centering Skateboard\end{minipage}
        \begin{minipage}{0.11\linewidth}\centering Table\end{minipage}

    \caption{Sample part segmentation results from ShapeNet-Part dataset.}
    \label{fig:shapenetparts}
\end{figure*}

\subsubsection{3D Object Classification}
We have performed the 3D object classification experiments on the ModelNet40~\cite{wu20153d} dataset, which is composed of 12,311 Computer Aided Design (CAD) models of 40 man-made object categories. 9,843 models are for training and 2,468 are for testing. In our experiments, following \cite{qi2017pointnet}, we used the pre-processed version of the dataset, which contains point clouds of objects, which are sampled from the original mesh models with cardinality 2048. The performance of our method and different baseline models are shown in Table~\ref{tab:modelnet}. It can be seen from the table that our proposed DPFA-Net is one of the top performers, and its overall accuracy is only slightly lower than that of GDANet, which indicates the ability of DPFA-Net for learning object representations. Moreover, our proposed DPFA-Net runs about six times faster that GDANet, and thus has a significant advantage in terms of computational efficiency as will be discussed in Section~\ref{ssec:performance-analysis}. It should also be noted that ModelNet40~\cite{wu20153d} dataset for 3D object classification is like MNIST dataset for 2D classification in that recently reported performances are close to each other.

\begin{table}[h!]
    \centering
    \caption{Classification results on ModelNet40\cite{wu20153d}.}
    \begin{tabular}{|l|c|}\hline 
    Model & OA(\%)\\\hline  
    DGCNN\cite{wang2019dynamic}& 92.9\\\hline 
    A-CNN\cite{komarichev2019cnn} & 92.6\\\hline
    PointWeb\cite{zhao2019pointweb} & 92.3\\\hline
    PointNet\cite{qi2017pointnet} & 89.2\\\hline
    PointNet++\cite{qi2017pointnet++} & 91.9\\\hline
    GDANet \cite{xu2020learning} & \textbf{93.8}\\\hline
    DPFA(ours) & 93.1\\\hline
    \end{tabular}
    \label{tab:modelnet}
\end{table}

\subsubsection{Object Part Segmentation}
The goal of object part segmentation is to effectively segment the 3D point clouds of objects into their parts. This is a particularly important yet challenging task for many applications. For example in robotics, the agents need to be able to recognize the parts of the object with which they interact. Thus, we have evaluated the effectiveness of our proposed DPFA-Net for part segmentation as well.

The part segmentation experiments were performed on the ShapeNet-Part dataset~\cite{yi2016scalable}, which is a common benchmark for part segmentation methods. It consists of densely (point-wise) annotated objects belonging to 16 categories. Each category has varying number of parts. For example, while the `knife' category has only 2 parts (sharp side and handle), the `motorbike' category has six parts corresponding to different portions of a motorbike. The ShapeNet-Part dataset contains 12137, 1870, and 2874 objects in its training, validation, and test sets, respectively. Table~\ref{tab:partsegmentation-table} shows the part segmentation results obtained with DPFA-Net and various baselines. Our proposed model provides the second best performance in terms of the mean shape IoU (Intersection-over-Union) score, and provides the top-3 performance for 12 out of the 16 individual object categories. Fig.~\ref{fig:shapenetparts} shows example qualitative results from each category. 

Overall, our proposed method provides a desirable trade-off by providing a consistently satisfactory performance across different tasks of semantic segmentation, part segmentation and 3D object classification, achieving either the best, or the second-best performance on different tasks, and also being computationally more efficient compared to other methods.


\subsection{Computational Efficiency Analysis}
\label{ssec:performance-analysis}
We have also performed some comparisons in terms of the computation times of different approaches. As presented above, our proposed method achieves the best overall accuracy on both Area-5 test and 6-fold cross validation on the semantic segmentation task. Even though RandLA-Net has a better mIoU score for this task, our model (with or without BF-Regularization) is approximately three times faster than RandLA-Net. More specifically, the proposed DPFA takes 28.88 ms and 29.58 ms, without and with BF-Regularization respectively, to process a single input from the S3DIS dataset, while the RandLA-Net takes 88.25 ms to process the same data. Similarly, for the classification task, even though the accuracy of our model is slightly lower than that of GDANet, our model is about six times faster than GDANet. For each input from ModelNet40, the average point processing time of our model is 4.43ms compared to the 27.46ms of GDANet. 

\section{Conclusion}\label{sec:conclusion}

In this paper, we have proposed a novel, deep 3D point cloud learning network by selectively performing the neighborhood feature aggregation with dynamic pooling and an attention mechanism. Our proposed Dynamic Point Feature Aggregation Network (DPFA-Net) has two variants for semantic segmentation and classification of 3D point clouds. As the core module of the DPFA-Net, we have proposed a Feature Aggregation layer, in which the features of the dynamic neighborhood of each point is aggregated via a self-attention mechanism to focus more on the relevant features in a local neighborhood. Our experimental results have shown that the proposed DPFA-Net achieves the state-of-the-art overall accuracy score on the S3DIS~\cite{armeni2017joint} dataset. Moreover, to further improve the performance of the proposed semantic segmentation model, we have proposed two novel approaches, namely Two-Stage BF-Net and BF-Regularization, to exploit the Background-Foreground information. They are designed so that they can be used with different existing semantic segmentation models on a plug-and-play basis. Experiments have demonstrated that these proposed BF exploitation techniques can improve the performance of the semantic segmentation approaches. We have also shown that our proposed approach provides a consistently satisfactory performance across different tasks of semantic segmentation, part segmentation and 3D object classification, and is also computationally more efficient compared to other methods.

\section*{Acknowledgment}

The information, data, or work presented herein was funded in part by National Science Foundation (NSF) under Grant 1739748, Grant 1816732 and by the Advanced Research Projects Agency-Energy (ARPA-E), U.S. Department of Energy, under Award Number DE-AR0000940. The views and opinions of authors expressed herein do not necessarily state or reflect those of the United States Government or any agency thereof. Authors would also like to thank HTC Campus Grid at Syracuse University for providing GPU resources under the NSF award ACI-1341006.

\bibliographystyle{IEEEtran}
\bibliography{paper}


\begin{IEEEbiography}[{\includegraphics[width=1in,height=1.25in,clip,keepaspectratio]{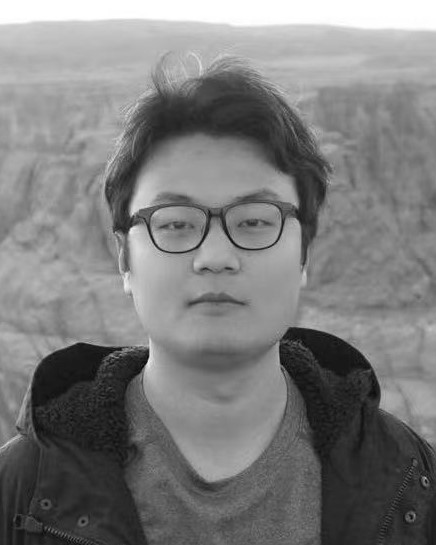}}]{Jiajing Chen}
received the B.S. degree in mechanical engineering from WuHan Institute of Technology, Wuhan, China in 2017 and M.S. degree in mechanical engineering from Syracuse University, Syracuse University, Syracuse, NY, USA in 2019. He is currently pursuing the Ph.D. degree with the Department of Electrical Engineering and Computer Science, Syracuse University. His research interests include Point Cloud Segmentation, Weakly Supervised Object Detection and Few-show Learning.
\end{IEEEbiography}

\begin{IEEEbiography}[{\includegraphics[width=1in,height=1.25in,clip,keepaspectratio]{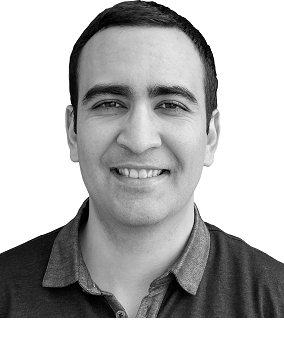}}]{Burak Kakillioglu}
received the B.S. degree in electrical and electronics engineering from Bilkent University, Ankara, Turkey, in 2015 and the Ph.D. degree in electrical and computer engineering from Syracuse University, Syracuse, NY, USA in 2021. He is currently a research scientist with the Motorola Solutions Inc., Somerville, MA, USA. His research interests include computer vision, machine intelligence, and embedded systems.
\end{IEEEbiography}

\begin{IEEEbiography}[{\includegraphics[width=1in,height=1.25in,clip,keepaspectratio]{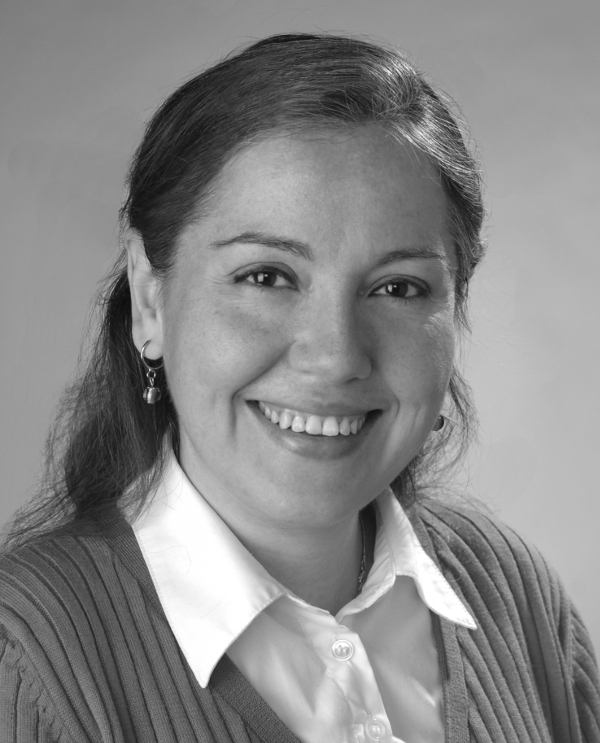}}]{Senem Velipasalar}

(M'04--SM'14) received the B.S. degree in electrical and electronic engineering from Bogazici University, Istanbul, Turkey, in 1999, the M.S. degree in electrical sciences and computer engineering from Brown University, Providence, RI, USA, in 2001, and the M.A. and Ph.D. degrees in electrical engineering from Princeton University, Princeton, NJ, USA, in 2004 and 2007, respectively. From 2001 to 2005, she was with the Exploratory Computer Vision Group, IBM T. J. Watson Research Center, NY, USA. From 2007 to 2011, she was an Assistant Professor with the Department of Electrical Engineering, University of Nebraska-Lincoln. She is currently a Professor in the Department of Electrical Engineering and Computer Science at Syracuse University.

The focus of her research has been on mobile camera applications, wireless embedded smart cameras, applications of machine learning, multi-camera tracking and surveillance systems, and automatic event detection from videos. Dr. Velipasalar received a Faculty Early Career Development Award (CAREER) from the National Science Foundation in 2011. She is the recipient of the 2014 Excellence in Graduate Education Faculty Recognition Award. She is the coauthor of the paper, which received the 2017 IEEE Green Communications and Computing Technical Committee Best Journal Paper Award. She received the Best Student Paper Award at the IEEE International Conference on Multimedia and Expo in 2006. She is the recipient of the EPSCoR First Award, IBM Patent Application Award, and Princeton and Brown University Graduate Fellowships. She is a member of the Editorial Board of the IEEE Transactions on Image Processing and Springer Journal of Signal Processing Systems.
\end{IEEEbiography}


\end{document}